\documentclass[fleqn,11pt,twocolumn]{wlscirep}
\usepackage[utf8]{inputenc}
\usepackage[T1]{fontenc}
\usepackage{graphicx,amssymb,amsmath,bm}
\usepackage[most]{tcolorbox}
\usepackage{multicol}
\usepackage{tabularx}
\usepackage{array}
\newcolumntype{C}[1]{>{\centering\arraybackslash}p{#1}}
\newcolumntype{?}{!{\vrule width 1pt}}
\usepackage{booktabs}
\usepackage{subcaption}
\usepackage{caption}
\usepackage{algpseudocodex}
\usepackage{algorithm}

\usepackage[justification=justified]{caption}

\usepackage{newtxtext,newtxmath}

\title{Machine learning in and out of equilibrium}

\author[1,2,3]{Shishir Adhikari}
\author[4,7]{Alkan Kabak\c c\i o\u glu}
\author[5]{Alexander Strang}
\author[6,7]{Deniz Yuret}
\author[1,*]{Michael Hinczewski}
\affil[1]{Department of Physics, Case Western Reserve University, Cleveland, OH, U.S.A.}
\affil[2]{Department of Systems Biology, Harvard Medical School, Boston, MA, U.S.A.}
\affil[3]{Department of Data Science, Dana-Farber Cancer Institute, Boston, MA, U.S.A. }
\affil[4]{Department of Physics, Ko\c c University, Istanbul, Turkey}
\affil[5]{Department of Statistics, University of Chicago}
\affil[6]{Department of Computer Engineering, Ko\c c University, Istanbul, Turkey}
\affil[7]{Ko\c{c} University, KUIS AI Center, Istanbul, Turkey}

\affil[*]{mxh605@case.edu}

\begin{abstract}

The algorithms used to train neural networks, like stochastic gradient descent (SGD), have close parallels to natural processes that navigate a high-dimensional parameter space---for example protein folding or evolution. Our study uses a Fokker-Planck approach, adapted from statistical physics, to explore these parallels in a single, unified framework. We focus in particular on the stationary state of the system in the long-time limit, which in conventional SGD is out of equilibrium, exhibiting persistent currents in the space of network parameters.  As in its physical analogues, the current is associated with an entropy production rate for any given training trajectory.  The stationary distribution of these rates obeys the integral and detailed fluctuation theorems---nonequilibrium generalizations of the second law of thermodynamics.  We validate these relations in two numerical examples, a nonlinear regression network and MNIST digit classification.  While the fluctuation theorems are universal, there are other aspects of the stationary state that are highly sensitive to the training details.  Surprisingly, the effective loss landscape and diffusion matrix that determine the shape of the stationary distribution vary depending on the simple choice of minibatching done with or without replacement.  We can take advantage of this nonequilibrium sensitivity to engineer an equilibrium stationary state for a particular application: sampling from a posterior distribution of network weights in Bayesian machine learning.  We propose a new variation of stochastic gradient Langevin dynamics (SGLD) that harnesses without replacement minibatching.  In an example system where the posterior is exactly known, this SGWORLD algorithm outperforms SGLD, converging to the posterior orders of magnitude faster as a function of the learning rate.
\end{abstract}

\begin{document}

\flushbottom
\maketitle
\thispagestyle{empty}

\section{Introduction}

Over the last decade, machine learning based on deep neural networks has profoundly impacted a wide variety of fields, including image recognition, natural language processing, health care, finance, manufacturing, autonomous driving, physics, engineering, structural biology, and others~\cite{carleo2019machine,goodfellow2016deep, mehta2019high}.  Many of these applications involve training a network:  finding model parameters that minimize the discrepancy between the ground truth (known from a set of training data) and the predicted value output by the model.  The discrepancy (the so-called loss function) depends on the network parameters, which can number in the billions or higher for the most complex problems.  The training process then becomes a search for a minimum in a highly multidimensional landscape defined by the loss function~\cite{li2018visualizing}.  Stochastic gradient descent (SGD), along with a multitude of variant methods derived from it, is one of the most popular algorithms for doing this minimization~\cite{robbins1951stochastic}.  In SGD each step of the training involves an approximation of the loss function, typically by using a small random subset (minibatch) of the total training data available to calculate the gradient with respect to the network parameters.  The parameters are then updated in a downward direction defined by this approximate gradient.  The resulting training process can be interpreted as biased diffusion on the "true" loss landscape defined by the whole dataset---biased because steps moving down along the true gradient are more likely than those moving up, and diffusive because of the effective noise due to the minibatch selection.

Biased diffusion on complex, multidimensional landscapes is a common motif in nature.  Two of the most well-known examples come from biology:  the free energy landscape which proteins navigate during folding~\cite{Bryngelson1995,veitshans1997protein,dill2017protein} and the fitness landscape on which Darwinian evolution plays out~\cite{wright1932roles,mustonen2010fitness,stadler2002fitness,de2018utility}.  In the simplest models of both processes (a single amino-acid chain folding at constant temperature, the Wright-Fisher model of evolution with fixed mutation and selection~\cite{hofrichter2017information}) diffusion leads in the long time limit to an equilibrium stationary distribution of physical observables (chain configurations, fractions of genetic variants in the population).  This stationary distribution is characterized by an environmental parameter---temperature for proteins, effective population size for evolution---that controls the extent of the fluctuations.  And since the systems are in equilibrium, the fluctuations obey local detailed balance: observing a transition between two states is equally likely as observing its reverse, so there is no net flow of probability.

In the case of SGD, we can ask whether a similar outcome occurs for sufficiently long training.  If the network parameters eventually reach a stationary distribution, is it in equilibrium?  If not, what features of this nonequilibrium stationary state are universal (governed for example by the fluctuation theorems of stochastic thermodynamics~\cite{seifert2012stochastic}), and which ones depend on the details of the training algorithm?  In certain applications, the precise nature of the fluctuations may not seem crucial, since we are only interested in a single sample from the distribution, a set of network parameters with low loss that performs well on the specified task.  But in the case of Bayesian machine learning the shape of the distribution is paramount, since we would like to sample network weights from a given posterior~\cite{welling2011bayesian}.  Can we leverage the knowledge of the distribution resulting from SGD to build more efficient Bayesian algorithms?

The Fokker-Planck (FP) equation provides a versatile mathematical formalism for exploring these questions, and for understanding the analogies between SGD and natural processes.  In biology, FP has been widely applied as a model of both protein dynamics~\cite{best2011diffusion} and evolution~\cite{kimura1954stochastic, mustonen2010fitness}.  FP in machine learning goes back more than thirty years to the work of Radons {\it et al.}~\cite{radons1990fokker}, which first showed how SGD could be mapped onto an FP equation.  Subsequent work has used this approach to argue that the SGD stationary state is fundamentally nonequilibrium~\cite{hansen1993stochastic,chaudhari2018stochastic}.  More recently, Feng and Tu showed that this nonequilibrium distribution obeys an inverse variance-flatness relation: in contrast to the equilibrium case, there is less variance in the network parameters along directions where the loss landscape is flatter~\cite{feng2021inverse}.  SGD thus seems to favor flat minima in the loss landscape, since it will spend longer time in them relative to steep minima, where it can escape more easily.  This perhaps explains some of the success of SGD at finding generalizable solutions---networks that perform well on novel data not seen during training.

The nonequilibrium nature of SGD also makes it amenable to the tools of stochastic thermodynamics.  In the current work, we harness those tools in the context of an FP model for SGD, to develop a comprehensive framework for understanding the distribution of network parameters.  We show there is an effective entropy production that can be defined for any training trajectory, and that this entropy not only satisfies an analog of the second law of thermodynamics but also integral and detailed fluctuation theorems~\cite{seifert2012stochastic}, two of the most important relations in modern nonequilibrium physics.  While these are universal laws that constrain the SGD distribution, there are other features that are sensitive to seemingly innocuous details of the training---for example the choice of whether to choose a minibatch with replacement (which allows consecutive batches to share some of the same data points) or without replacement (which ensures that all points are sampled once during a training period known as an epoch).  The without replacement (WOR) case requires a careful formulation to ensure that it can be described via FP, since one must look at how the parameters change epoch-to-epoch in order to guarantee Markovian dynamics.  Surprisingly WOR leads to a very different noise distribution than the simpler with replacement (WR) case and an effective loss landscape that is perturbed from the true one.  The theory works for all network architectures, and we numerically validate its predictions by training on a simple regression problem (for illustrative clarity) as well as MNIST digit recognition.

Finally, we argue that knowing the nature of the nonequilibrium distribution allows us to more effectively engineer an equilibrium one. This forced equilibration is at the heart of stochastic gradient Langevin dynamics (SGLD)~\cite{welling2011bayesian}, a widely studied approach to Bayesian machine learning. SGLD injects additional noise into SGD training in order to approximately sample network weights from a desired posterior, which happens to be an equilibrium distribution.  Taking advantage of the fact that the spread of the distribution for WOR training is typically smaller than the WR case (when sampled at epoch intervals) we introduce a new algorithm:  stochastic gradient without replacement Langevin dynamics (SGWORLD).  In a test system where the posterior is exactly known, we show that SGWORLD converges orders of magnitude faster than SGLD toward the true posterior, as a function of the learning rate.

The organization of the paper is as follows:  Sec.~\ref{theory} provides a general overview of the FP formalism, how we use it to characterize stationary states, and its connections to stochastic thermodynamics.  We also contrast SGD with biased diffusion in biological contexts (Box 1).  Sec.~\ref{ness} delves into the specifics of nonequilibrium stationary state for SGD dynamics, and the unexpected distinctions between WR and WOR minibatching.  Sec.~\ref{results} covers the numerical verification of the theory and discusses the implications of our approach. We show that the training satisfies the integral and detailed fluctuation theorems.  The theory also provides insights into the origins of the inverse-flatness relation, by comparing SGD to an alternative stochastic gradient approach we call the earthquake model.  Both approaches exhibit the same inverse variance-flatness relation, but the distribution in the earthquake model is equilibrium (on a different loss landscape).  The last part of Sec.~\ref{results} describes the SGWORLD algorithm and demonstrates its efficiency.

\section{Theoretical Methods}\label{theory}

\subsection{Fokker-Planck equation and classification of stationary states}

We begin by reviewing the FP formalism that will be our main tool for understanding SGD dynamics and its parallels to other physical systems~\cite{risken1996fokker}.  Let us imagine we are interested in the stochastic dynamics of a continuous vector quantity $\bm{\theta}$, with components $\theta_\alpha$, $\alpha = 1,\ldots,N$.  We will always denote indices that run up to $N$ by Greek letters.  As we will see below, in our case these $\bm{\theta}$ correspond to the parameters of a neural network, but for now we will keep the discussion general.  Specific interpretations of the quantities in the FP formalism for the examples of SGD, protein folding, and evolution are summarized in Box 1.

The FP equation describes the time evolution of $P(\bm{\theta},t)$, the probability density of observing $\bm{\theta}$ at time $t$.  It takes the form of a continuity equation,
\begin{equation}
    \label{fp1}
    \frac{\partial}{\partial t}P(\bm{\theta},t) = -\bm{\nabla}\cdot \bm{\mathcal J}(\bm{\theta},t).
\end{equation}
Unless otherwise noted, all vector calculus operations (like divergences or gradients) are defined with respect to $\bm{\theta}$.  The probability current density $\bm{\mathcal J}(\bm{\theta},t)$ has the general form,
\begin{equation}
\label{fp2}
    \bm{\mathcal J}(\bm{\theta},t) = - P(\bm{\theta},t) \mu(\bm{\theta}) \bm{\nabla} \mathcal{L}(\bm{\theta}) - \bm{\nabla} \cdot \left[D(\bm{\theta}) P(\bm\theta,t) \right].
\end{equation}
The first term, known as the drift, describes an effective force $-\bm{\nabla} \mathcal{L}(\bm{\theta})$ that pushes opposite to the gradient of a potential function $\mathcal{L}(\bm{\theta})$.  The symmetric $N\times N$ matrix $\mu(\bm{\theta})$ is the mobility (or inverse friction) that scales how the force translates into current.

The second term describes diffusion, or the spreading of the probability distribution in $\bm{\theta}$ space due to stochastic effects.  $D(\bm{\theta})$ is a symmetric $N \times N$ diffusion matrix, and we use the convention that the divergence $\bm{\nabla} \cdot A(\bm{\theta})$ of a matrix $A(\bm{\theta})$ is a vector representing the divergence of each of the matrix's rows:  $(\bm{\nabla} \cdot A(\bm{\theta}))_\alpha = \sum_\gamma \partial_\gamma A_{\alpha\gamma}(\bm{\theta})$, where $\partial_\gamma \equiv \partial / \partial \theta_\gamma$.  $D(\bm\theta)$ is generically positive semi-definite, but unless otherwise noted we will restrict ourselves to the positive definite case (which we will show is sufficient to describe SGD dynamics).

The above framework, where the dynamics depend explicitly only on $\bm{\theta}$ and not its time derivative $\dot{\bm{\theta}}$, is known as overdamped FP.  This turns out to describe conventional SGD and the variants we consider here.  However the FP description is generalizable to training algorithms that include inertial terms (i.e. those involving momentum), and would lead to an underdamped FP for the joint distribution $P(\bm{\theta},\dot{\bm{\theta}},t)$~\cite{kunin2021rethinking}.

The nature of FP dynamics depends on the form of and relationships between the quantities ${\cal L}$, $\mu$, and $D$ that determine Eq.~\eqref{fp2}.  One way of classifying FP systems is to look at the stationary distribution $P^s(\bm{\theta})$ that is the long-time limit of $P(\bm{\theta},t)$.  The corresponding stationary current density $\bm{\mathcal J}^s(\bm{\theta})$, which is just Eq.~\eqref{fp2} with $P^s$ substituted for $P$, satisfies $\bm{\nabla}\cdot \bm{\mathcal J}^s(\bm{\theta}) = 0$.  Stationary states are in {\it equilibrium} if $\bm{\mathcal J}^s(\bm{\theta})=0$, and {\it non-equilbrium} if $\bm{\mathcal J}^s(\bm{\theta})\ne0$.

To understand this distinction better, let us write the stationary distribution in the form $P^s(\bm{\theta}) = \exp(-\Phi(\bm{\theta}))$, which defines a pseudopotential $\Phi(\bm{\theta})$.  The stationary current density can then be written as $\bm{\mathcal J}^s(\bm\theta) = \bm{v}^s(\bm\theta) P^s(\bm{\theta})$, where $\bm{v}^s(\bm\theta)$ is a velocity field given by:
\begin{equation}
    \label{fp2b}
    \bm{v}^s(\bm\theta) = D(\bm{\theta}) \bm{\nabla} \Phi(\bm\theta) - \mu(\bm\theta) \bm{\nabla} {\mathcal L}(\bm\theta) - \bm{\nabla} \cdot D(\bm\theta).
\end{equation}
Equilibrium corresponds to $\bm{v}^s(\bm\theta) = \bm{0}$.  There are two special cases of stationary states found in physical systems that are relevant to our discussion:

{\bf 1) {\it\bf Generalized Boltzmann equilibrium (GBE)}}. This occurs when two conditions are satisfied: 
\begin{enumerate}
    \item[(i)] Einstein relation: the matrices $\mu$ and $D$ are related through $\mu(\bm{\theta}) = \beta D(\bm{\theta})$, with a proportionality constant $\beta$;
    \item[(ii)] Divergence condition:  there exists a scalar function $G(\bm\theta)$ such that $\bm{\nabla}\cdot (G(\bm\theta) D(\bm\theta)) = 0$.
\end{enumerate}
Given these conditions, we see from Eq.~\eqref{fp2b} that $\bm{v}^s(\bm\theta) = \bm{0}$ when $\Phi(\bm\theta) = \beta {\mathcal L}(\bm\theta) - \log G(\bm\theta)$, up to a constant, and hence
\begin{equation}
    \label{fp3}
    P^s(\bm{\theta}) = Z^{-1} G(\bm\theta)\exp(-\beta {\cal L}(\bm{\theta})),
\end{equation}
where $Z$ is a normalization constant ensuring that $\int d\bm{\theta}\, P^s(\bm\theta) = 1$.  

GBE reduces to conventional Boltzmann equilibrium (BE) when $G(\bm\theta)$ is a constant, as is the case for protein folding and many other thermodynamic systems that reach equilibrium (see Box~1).  Here $\beta \propto T^{-1}$, where $T$ is the temperature of the environment.  For evolution, in contrast, $G(\bm\theta)$ is non-constant, and $\beta \propto S^{-1}$, where $S$ is the effective population size.

As we will see below when we discuss thermalized SGD, it turns out condition (i) is the critical element of getting a system to exhibit GBE.  Condition (ii) can always be engineered through a coordinate transformation, if necessary.  To understand this, we can use a geometrical interpretation: when $D(\bm\theta)$ is a positive-definite matrix, then the FP equation can be equivalently written in a covariant form, and interpreted as describing diffusion on a Riemannian manifold~\cite{graham1977covariant}.  The metric tensor for this manifold in the $\bm\theta$ coordinates is just the inverse of the $D(\bm\theta)$ matrix.  Once the metric is known, the curvature of the manifold can then be calculated using the standard tools of Riemannian geometry.  From this viewpoint, for example, the evolution FP summarized in Box 1 corresponds to diffusion on a spherical manifold.  A $D(\bm\theta)$ matrix independent of $\bm\theta$ would correspond to a flat manifold.  

Condition (ii) is equivalent to a gauge condition defining a coordinate system on the manifold.  In some cases (like protein folding or evolution) the standard coordinates used to model the system already satisfy the condition.  In other cases, one can enforce the condition by transforming to new coordinates $\bm\theta^\prime(\bm\theta)$ that are functions of the original $\bm\theta$.  This is routinely done in general relativity calculations, where enforcing condition (ii) with $G(\bm\theta^\prime) = (\det D(\bm\theta^\prime))^{-1/2}$ defines so-called ``harmonic coordinates'' that simplify certain equations.  In general, coordinates that satisfy condition (ii) are guaranteed to exist locally in the vicinity of any point on the manifold~\cite{wald2010general}.  In practice they may extend to nearly the whole manifold, though in certain systems a single coordinate chart cannot cover the entirety of the manifold (i.e. leaving out the north pole in the case of a sphere).  

{\bf 2) {\it\bf Nonequilibrium stationary state (NESS)}}.  This occurs when $\bm{v}^s(\bm\theta) \ne 0$, and implies that condition (i) is violated (since condition (ii) can always be enforced via a coordinate transformation if necessary).  As described below, conventional SGD tends to this type of stationary state.  

\subsection{Trajectory picture and properties of a non-equilibrium stationary state}\label{trajpic}

In order to understand the nature of the NESS in more detail, it is useful to consider individual stochastic trajectories of $\bm\theta(t)$, along with the distribution $P(\bm\theta,t)$.  Stochastic trajectories compatible with the Fokker-Planck system of Eqs.~\ref{fp1}-\ref{fp2} satisfy the following It\^{o} stochastic differential equation(SDE)~\cite{gardiner1985handbook}:
\begin{equation}\label{tr1}
d\bm\theta(t) = -\mu(\bm\theta(t)) \bm\nabla {\cal L}(\bm\theta(t)) dt + Z(\bm\theta(t)) d\bm{W}(t),
\end{equation}
where $d\bm{W}(t)$ is an $N$-dimensional Wiener process, and $Z(\bm{\theta})$ is an $N \times N$ matrix that satisfies $\frac{1}{2} Z(\bm\theta) Z^T(\bm\theta) = D(\bm\theta)$.  Such a decomposition always exists for positive semi-definite $D(\bm\theta)$.

Let us imagine we have let the system reach a stationary state, and follow an individual stochastic trajectory $\Omega \equiv \{\bm\theta(t) | t_0 \le t \le t_0 + \tau\}$ over time $\tau$.  We can define two quantities associated with this trajectory.  The first is an $N\times N$ stochastic area matrix~\cite{ghanta2017fluctuation} $A_{\alpha\gamma}(\Omega)$, where each element is given by a line integral over $\Omega$:
\begin{equation}
    \label{tr2}
    A_{\alpha\gamma}(\Omega) = \frac{1}{2}\int_\Omega \left[(\theta_\alpha-\theta_{0\alpha}) d\theta_\gamma - (\theta_\gamma-\theta_{0\gamma}) d\theta_\alpha\right].
\end{equation}
Here $\bm\theta_0$ is an arbitrary reference point, which we will typically take to be the location of a minimum in ${\cal L}(\bm\theta)$.  The element $A_{\alpha\gamma}(\Omega)$ can be interpreted as the oriented area in the $\theta_\alpha$, $\theta_\gamma$ plane swept out by the trajectory relative to the origin at $\bm\theta_0$.  Every infinitesimal counterclockwise contribution to the area contributes positively to the integral, while clockwise contributions are negative.  Intuitively, we expect the non-vanishing current of an NESS to be associated with a mean overall circulation of the probability.  Let us define the average $\langle A_{\alpha\gamma} \rangle$ as taken over the ensemble of all NESS trajectories with a certain length $\tau$.  This average will depend on the $\tau$ defining the ensemble, and if there is a nonzero mean circulation the area swept out will grow with longer trajectories: $\langle A_{\alpha\gamma} \rangle \to \pm \infty$ as $\tau \to \infty$.  We will indeed find for our conventional SGD case that $\langle A_{\alpha\gamma} \rangle / \tau$ converges to a non-zero value for large $\tau$.

The second quantity is based on an analogy from stochastic thermodynamics~\cite{seifert2012stochastic}.  We can define an entropy production $\sigma(\Omega)$ via the line integral
\begin{equation}
    \label{tr3}
    \sigma(\Omega) = \int_{\Omega} \bm{v}^s(\bm\theta) \cdot D^{-1}(\bm\theta) d\bm\theta,
\end{equation}
where $\bm{v}^s(\bm\theta)$ is the NESS velocity field given in Eq.~\eqref{fp2b}.  This quantity obeys the analog of the second law of thermodynamics, in the sense that its ensemble average is always non-negative, $\langle \sigma \rangle \ge 0$, and is strictly greater than zero for an NESS.  Moreover, the entropy production also satisfies other interesting relations that have become mainstays of nonequilibrium thermodynamics~\cite{seifert2012stochastic}:  i) the {\it integral fluctuation theorem} (IFT) given by
\begin{equation}
    \label{tr4}
    \langle \exp(-\sigma) \rangle =1,
\end{equation}
and the {\it detailed fluctuation theorem} (DFT),
\begin{equation}
    \label{tr5}
    \frac{{\cal P}(-\sigma)}{{\cal P}(\sigma)} =  \exp(-\sigma).
\end{equation}
Here ${\cal P}(\sigma)$ is the probability density of observing a trajectory with entropy production value $\sigma$ in the ensemble.  As in the case of the average area matrix, the DFT reflects a breaking of symmetry in the NESS:  trajectories with negative entropy production are less likely in the ensemble than those with positive entropy.  In fact the DFT implies the IFT, which in turn implies the second law.  Additionally, as we highlight below, the ensemble averages of the entropy production $\langle \sigma \rangle$ and the area matrix $\langle A_{\alpha\gamma} \rangle$ can be directly related to each other~\cite{tomita2008irreversible}. All the above properties of NESS trajectories will manifest themselves in our analysis of conventional SGD.  

\begin{figure*}
\begin{tcolorbox}[
  colback=white,
  title={~\centering Box 1: Comparing stochastic search dynamics in protein folding, evolution, and machine learning}]

\centering\includegraphics[width=0.9\textwidth]{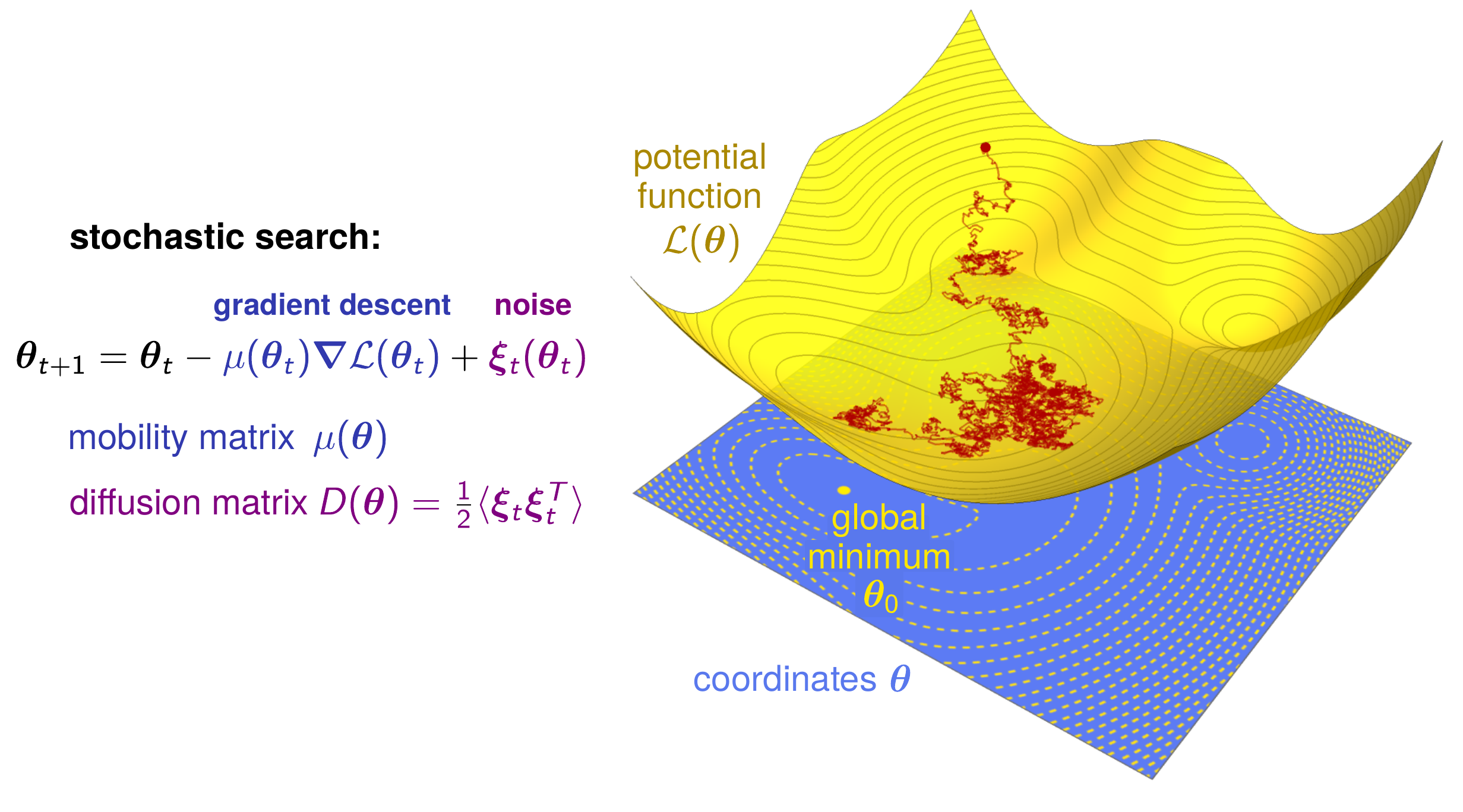}

\bgroup

\begin{tabular}{C{0.7in}p{1.7in}p{1.71in}p{1.7in}}
\toprule
&&&\\[-1em]
 & {\bf Protein folding} & {\bf Evolution} & {\bf Machine learning}\\
 & polymer in solution & Wright-Fisher model & conventional SGD training\\
 &&&\\[-1em]
 \toprule
 &&&\\[-1em]
 {\bf coordinates} $\bm{\theta}$ & monomer positions & fraction of each genetic variant in the population & neural network weights\\
&&&\\[-1em]
\midrule
&&&\\[-1em]
{\bf potential} ${\cal L}(\bm{\theta})$ & free energy describing interactions between monomers & fitness landscape function describing the effects of natural selection and mutation & loss function, with additional corrections if minibatching without replacement\\
&&&\\[-1em]
\midrule
&&&\\[-1em]
{\bf mobility matrix} $\mu(\bm\theta)$& diagonal elements of $\mu(\bm\theta)$:  inverse friction coefficient of each monomer; off-diagonal elements of $\mu(\bm\theta)$: non-local effects due to hydrodynamics & matrix $\mu(\bm\theta)$ encodes the constraint that the fractions of all the variants must sum to one & $\mu(\bm\theta) = \eta I$, where $\eta$ is the learning rate and $I$ the identity matrix\\
&&&\\[-1em]
\midrule
&&&\\[-1em]
{\bf diffusion matrix} $D(\bm\theta)$ & $\text{Einstein relation holds:}$\hspace{0.5in} $D(\bm\theta) = \beta^{-1} \mu(\bm\theta)$, where $\beta \propto T^{-1}$ and $T$ is the temperature& $\text{Einstein relation holds:}$\hspace{0.5in} $D(\bm\theta) = \beta^{-1} \mu(\bm\theta)$, where $\beta \propto S$, the effective population size& Einstein relation does not hold:  $D(\bm\theta)$ is determined by the form of the loss function and details of the minibatching procedure\\
&&&\\[-1em]
\midrule
&&&\\[-1em]
{\bf divergence condition} & $\bm\nabla \cdot (G(\bm\theta) D(\bm\theta)) = 0$ holds with $G(\bm\theta) = 1$ & $\bm\nabla \cdot (G(\bm\theta) D(\bm\theta)) = 0$ holds with $G(\bm\theta) = (\det D(\bm\theta))^{-1}$& $\bm\nabla \cdot (G(\bm\theta^\prime) D(\bm\theta^\prime)) = 0$ can be enforced via a coordinate transformation, $\bm\theta \to \bm\theta^\prime$\\ 
&&&\\[-1em]
\midrule
&&&\\[-1em]
{\bf stationary state} $P^s(\bm\theta)$ & $\text{Boltzmann equilibrium:}$ $P^s(\bm\theta) \propto \exp(-\beta {\cal L}(\bm\theta))$ & $\text{generalized Boltzmann}$ $\text{equilibrium:}$ \hspace{0.5in}$P^s(\bm\theta) \propto G(\bm\theta)\exp(-\beta {\cal L}(\bm\theta))$ & nonequilibrium stationary state\\
\addlinespace[0.5em]
\bottomrule
\end{tabular}
\egroup
\label{box1}
\end{tcolorbox}
\end{figure*}

\section{Conventional SGD dynamics lead to a nonequilibrium stationary state}\label{ness}

In this section, we explore how the training dynamics of conventional SGD fit into the theoretical framework outlined above.  The precise nature of the stationary state achieved by the dynamics depends on the details of minibatching procedure, and hence we will discuss different types of minibatching separately.  We begin with the simpler case of minibatching with replacement (WR).

\subsection{SGD based on a minibatching with replacement}\label{sec:sgd_wr}

Consider a neural network whose $N$ parameters $\theta_\alpha$, $\alpha = 1,\ldots,N$ are trained over a data set of $M$ input-output pairs $\bm{z}_i \equiv (\bm{x}_i, \bm{y}_i)$,  $i=1,\ldots,M$, where $\bm{x}_i$ is an input vector and $\bm{y}_i$ is the corresponding desired output vector.  We will denote indices that run up to $M$ by Roman letters.  The training process is a search over the $\bm\theta$ parameter space to minimize a scalar loss function ${\cal L}(\bm{\theta})$ that encodes the performance of the network in generating the correct output for each input vector.  When the number of model parameters is large ($N \gg M$), the search is likely to yield an overfitting model which does not generalize well. Such an outcome can be avoided by including a suitable regularization, or constraint function, in ${\cal L}(\bm\theta)$ and/or using other techniques, such as, early termination or dropout~\cite{srivastava2014dropout,goodfellow2016deep}. We focus here on the regularized scenario, where the overall loss function ${\cal L}(\bm{\theta})$ can be expressed in two parts:
\begin{equation}
    \label{sgd1}
    {\cal L}(\bm\theta) = \frac{1}{M}\sum_{i=1}^M \ell( \bm{z}_i;\bm{\theta}) + R(\bm\theta) \equiv L(\bm\theta) + R(\bm\theta).
\end{equation}
Here $\ell(\bm\theta;\bm{z}_i)$ represents the additive contribution of the $i$th training datapoint to the loss, and $R(\bm\theta)$ is a constraint function, independent of the training data, that acts as regularization.  For example, if the network output, given a set of parameters $\bm\theta$ and input $\bm{x}_i$, is $\bm{f}(\bm{x}_i;\bm\theta)$ then a mean squared loss function would be defined via $\ell(\bm{z}_i;\bm\theta) = (\bm{y}_i - \bm{f}(\bm{x}_i;\bm\theta))^2$.  $L_2$ regularization would look like $R(\bm\theta) = \lambda {\bm\theta}^2$ for some constant $\lambda$.  For now, our discussion is general and holds for any specific forms of $\ell(\bm{z}_i;\bm\theta)$ and $R(\bm\theta)$. 

The parameters corresponding to a local minimum of ${\cal L}$ can be reached by gradient descent:
\begin{equation}
\label{eq_gd}
\bm{\theta}_{t+1} = \bm{\theta}_{t} - \eta\,  \bm{\nabla} {\cal L}(\bm{\theta}_t),
\end{equation}
where ${\bm\theta}_t$ are the network parameters at the $t$th iteration step, and $\eta$ is the learning rate, typically chosen to be $\ll 1$.  The number of training examples is usually large ($M\gg 1$), making the gradient calculation at each iteration of Eq.~\eqref{eq_gd} computationally costly. As a remedy in many practical applications, $L(\bm{\theta}_t)$ is approximated by $L_{B_t}(\bm{\theta}_t) = m^{-1}\sum_{i \in B_t} \ell(\bm{\theta}_t;\bm{z}_i)$, where $B_t$ is a set of $m$ distinct integers drawn from the range 1 to $M$.  The training examples corresponding to the labels in $B_t$ are known as a minibatch of size $m$, and we usually choose $m \ll M$ to reduce the computational burden at each iteration step (the limit $m=M$ would correspond to the regular gradient descent of Eq.~\eqref{eq_gd}).  WR minibatching means that the set $B_t$ is drawn independently at each step $t$, with the possibility that the same examples appear in consecutive minibatches.  When $m < M$ we have stochastic gradient descent (SGD), which can be expressed as a perturbation to Eq.~\eqref{eq_gd},
\begin{equation}
\label{eq_sgd}
\bm{\theta}_{t+1} = \bm{\theta}_{t} - \eta\,  \bm{\nabla} {\cal L}(\bm{\theta}_t) + \bm{\xi}_t.
\end{equation}
Here the vector $\bm{\xi}_t \equiv \eta\,\bm{\nabla} \big[L(\bm{\theta}_t) -L_{B_t}(\bm{\theta}_t)\big]$ is a stochastic quantity, dependent on the random choice of $B_t$.  
When $M \gg m \gg 1$ the distribution of $\bm{\xi}_t$ for the ensemble of all possible $B_t$ at each $t$ is approximately Gaussian with zero mean due to the central limit theorem, an assumption which we will later validate using our numerical examples.  This allows us to interpret Eq.~\eqref{eq_sgd} as an Euler method approximation to the It\^{o} SDE of Eq.~\eqref{tr1}, with time units chosen so that the time increment of each step $\Delta t =1$.  The corresponding Fokker-Planck equation has mobility and diffusion matrices given by:
\begin{equation}
    \label{sgd2}
    \mu(\bm{\theta}) = \eta I, \quad D(\bm\theta) = \frac{1}{2} \langle \bm{\xi}_t \bm{\xi}_t^T \rangle,
\end{equation}
where $I$ is the $N$-dimensional identity matrix and the $\langle \: \rangle$ brackets denote an average over all possible minibatches $B_t$.  As shown in the Appendix A, this average can be calculated analytically, giving
\begin{equation}
    \label{sgd3} 
    \begin{split}
    D(\bm\theta)&=\frac{\eta^2(n-1)}{2(M-1)}\left[MV(\bm\theta)V^T(\bm\theta) - \bm{\nabla}L(\bm\theta)\bm{\nabla}L^T(\bm\theta)\right]\\
    &\approx \frac{\eta^2 n}{2}V(\bm\theta)V^T(\bm\theta),
    \end{split}
\end{equation}
where $n=M/m$ and $V(\bm\theta)$ is an $N\times M$ dimensional matrix with components $V_{\alpha i}(\bm{\theta}) = M^{-1}\partial_\alpha \ell(\bm{z}_i;\bm{\theta})$, with $\partial_\alpha \equiv \partial/\partial \theta_\alpha$.  The second line is the approximate form in the limit where $M\gg 1$ and $n \gg 1$.  As expected, when the minibatch size $m$ increases and hence $n$ decreases, the stochasticity of SGD is diminished, leading to a smaller magnitude of $D(\bm\theta)$.  From the expressions for $\mu(\bm\theta)$ and $D(\bm\theta)$ in Eqs.~\eqref{sgd2}-\eqref{sgd3}, we see that the Einstein relation, $\mu(\bm\theta) \propto D(\bm\theta)$, will not be true in general.  Thus the system should exhibit an NESS, whose properties we will return to below.  However even without knowing the explicit form for the stationary distribution $P^s(\bm\theta)$, Eqs.~\eqref{sgd2}-\eqref{sgd3} already tell us something about its scaling behavior.  If the learning rate and batch size are scaled by the same factor, $\eta \to c \eta$, $m \to c m$, this scales the mobility and diffusion matrices identically:  $\mu(\bm\theta) \to c \mu(\bm\theta)$, $D(\bm\theta) \to c D(\bm\theta)$.  From Eq.~\eqref{fp2} this means the current $\bm{\mathcal J}(\bm{\theta},t)$ also gets scaled by a factor of $c$.  As a result any stationary state solution for the original system, which satisfies $\bm{\nabla}\cdot \bm{\mathcal J}^s(\bm{\theta}) = 0$, will also be a stationary state for the rescaled system.  Within the limits of the approximation ($M\gg 1$ and $n \gg 1$) we thus recover an important scaling property of the SGD stationary state that has been pointed out in previous work~\cite{jastrzkebski2017three}.  

There may be scenarios with certain datasets where a subset of network parameters $\theta_\alpha$ do not affect the loss function, $\partial_\alpha \ell(\bm{z}_i;\bm\theta) = 0$ for all $i$, and hence are not trainable (they remain unaltered under SGD).  For each such $\alpha$, the corresponding rows and columns of $D(\bm\theta)$ from Eq.~\eqref{sgd3} will be zero.  $D(\bm\theta)$ will then be positive semi-definite rather than positive definite.  To avoid this complication, we can always focus on the trainable parameters, which are the only ones relevant for describing SGD dynamics.  Thus we will assume $D(\bm\theta)$ is positive definite.

\subsection{SGD based on a minibatching without replacement}\label{sec:sgd_wor}

The potential landscape and diffusion characteristics of SGD depend on the details of the training algorithm.  Even a seemingly simple modification like doing the minibatching without replacement (WOR) can have non-trivial consequences.  In this case we are guaranteed to sample all $M$ data points every $n=M/m$ time steps.  Within a sequence of $n$ consecutive time steps, which we call an epoch, the sets $B_t$ (each of size $m$) are not independent of one another, since every integer from 1 to $M$ only appears in one $B_t$ per epoch.  This introduces correlations between time steps in an epoch, and hence a non-Markovian dependence of the noise $\bm{\xi}_t$ in Eq.~\eqref{eq_sgd} on the past history $\bm{\xi}_{t^\prime}$ for $t^\prime < t$ within the same epoch.  However since the random sampling is done independently in different epochs, there is a way to construct a Markovian analogue of Eq.~\eqref{eq_sgd} by considering the changes in network parameters from epoch to epoch.  Defining $\hat{\bm{\theta}}_\tau \equiv \bm\theta_{\tau n}$ as the parameters after the $\tau$th epoch, we show in Appendix B that the update equation takes the form
\begin{equation}
    \label{sgd_wor}
    \bm{\hat\theta}_{\tau+1} = \bm{\hat\theta}_{\tau} - \eta\,  \bm{\nabla} \hat{\cal L}(\bm{\hat\theta}_\tau) + \bm{\hat\xi}_\tau,
\end{equation}
where the new zero-mean Gaussian noise term $\bm{\hat\xi}_\tau$ is drawn independently from epoch to epoch.  Interestingly, the effective loss landscape $\hat{\cal L}(\bm{\theta})$ is a scaled version of the original $\cal{L}(\bm\theta)$ with a small perturbation, here shown to first order in $\eta$:
\begin{equation}
    \label{sgd4}
    \begin{split}
    \hat{\cal L}(\bm{\theta}) &= n {\cal L}(\bm{\theta}) - \frac{\eta n(n-1)}{4} \Biggl[ \left|\bm\nabla {\cal L}(\bm{\theta})\right|^2 + \frac{|\bm\nabla L(\bm{\theta})|^2}{M-1}\\
    & \qquad -\frac{M}{M-1}\text{tr}\left(V(\bm{\theta})V^T(\bm{\theta}) \right)\Biggr] + {\cal O}(\eta^2)\\
    &\equiv n {\cal L}(\bm{\theta}) + \delta\!{\cal L}(\bm{\theta}).
    \end{split}
\end{equation}
The perturbation $\delta\!{\cal L}(\bm{\theta})$ means that a local minimum for ${\cal L}$, where $\bm{\nabla} {\cal L}(\bm\theta) =0$, will no longer be a local minimum for $\hat{\cal L}$.  We will see this displacement in the minimum directly in our numerical examples below.

Relating Eq.~\eqref{sgd_wor} to the corresponding Fokker-Planck system, the analogue of Eq.~\eqref{sgd2} is:
\begin{equation}
    \label{sgd6}
    \hat\mu(\bm{\theta}) = \eta I, \quad \hat D(\bm{\theta}) = \frac{1}{2} \langle \bm{\hat\xi}_\tau \bm{\hat\xi}_\tau^T \rangle,
\end{equation}
and we can again evaluate $\hat D(\bm{\theta})$ explicitly.  The resulting expression, which is far more complex than Eq.~\eqref{sgd3}, can be found in Appendix B.  However, there is a simple approximation in the $M \gg 1$, $n \gg 1$ limit that works well in all the cases we have tested,
\begin{equation}
\label{sgd7}
\begin{split}
    \hat D(\bm{\theta}) &\approx \frac{(\eta n)^4}{24}H(\bm\theta)V(\bm\theta)V^T(\bm\theta)H(\bm\theta)\\
                        &\approx \frac{\eta^2 n^3}{12} H(\bm\theta) D(\bm\theta) H(\bm\theta).
\end{split}
\end{equation}
Here $H(\bm\theta)$ is the symmetric Hessian matrix for the original loss landscape, with components $H_{\alpha\gamma}(\bm\theta) = \partial_\alpha \partial_\gamma {\cal L}(\bm\theta)$ that describe the landscape curvature.  In the second line we have expressed $\hat D(\bm{\theta})$ in terms of the earlier diffusion matrix $D(\bm{\theta})$ from Eq.~\eqref{sgd3}.  We see that switching from WR to WOR minibatching alters the diffusion characteristics in a way that depends on the curvature of the loss landscape.  What remains the same is the fact the Einstein relation is violated, leading to an NESS in both scenarios, albeit with different distributions $P^s(\bm\theta)$.  

\subsection{Properties of the SGD nonequilibrium stationary state}\label{sec:ss}

Many of the characteristics of the NESS for conventional SGD can be derived analytically in the limit of small learning rate $\eta \ll 1$.  To be concrete, we will focus on the case of WR minibatching.  However the WOR case can be treated using the same formalism, substituting $\hat{\cal L}$ for ${\cal L}$ and $\hat D$ for $D$ in the discussion below.

We assume that after sufficient training steps we have arrived in the vicinity of local minimum $\bm\theta_0$ of ${\cal L}(\bm\theta)$, where $\bm\nabla {\cal L}(\bm\theta_0) = 0$.  Because $\eta$ is small, the stationary state consists of fluctuations around $\bm\theta_0$, in a loss landscape whose local curvature is approximately described by the Hessian $H_0 = H(\bm\theta_0)$.  The resulting local gradient can be written as $\bm\nabla {\cal L}(\bm\theta) \approx H_0 \delta \bm{\theta}$, where $\delta\bm\theta \equiv \bm\theta - \bm\theta_0$.  Similarly we assume $D(\bm\theta)$ does not change significantly near $\bm\theta_0$, so we can approximate it as $D_0 = D(\bm\theta_0)$.

With these assumptions the Fokker-Planck current density from Eq.~\eqref{fp2} simplifies to
\begin{equation}
\label{n1}
    \bm{\mathcal J}(\bm{\theta},t) \approx -\eta P(\bm{\theta},t) H_0 \delta\bm\theta - D_0 \bm{\nabla} P(\bm\theta,t),
\end{equation}
where we substituted $\mu(\bm\theta) = \eta I$ from Eq.~\eqref{sgd2}.  The stationary solution $P^s(\bm{\theta}) = \exp(-\Phi(\bm{\theta}))$ for this linearized Fokker-Planck system is analytically solvable~\cite{tomita1974irreversible}.  Here we give an summary of the main results, and provide a detailed derivation in Appendix C.  The pseudopotential is quadratic,
\begin{equation}\label{n2}
    \Phi(\bm\theta) = \Phi_0 + \frac{1}{2}\delta\bm\theta^T \Sigma^{-1} \delta\bm\theta,
\end{equation}
where $\Phi_0$ is a constant and $\Sigma$ is the stationary covariance matrix, $\Sigma = \int d\bm\theta P^s(\bm{\theta}) \delta\bm\theta \delta\bm\theta^T = \langle \delta\bm\theta \delta \bm\theta^T\rangle$.  This symmetric matrix $\Sigma$ satisfies a Lyapunov relation,
\begin{equation}\label{n3}
    H_0 \Sigma +\Sigma H_0 = 2\eta^{-1} D_0,
\end{equation}
which plays the role of a generalized fluctuation-dissipation theorem:  it relates the magnitude of the stationary fluctuations, characterized by $\Sigma$, to the diffusion matrix $D_0$.  In fact taking the trace of both sides of Eq.~\eqref{n3} and dividing by 2, we find
\begin{equation}\label{n3b}
\text{tr}\, \langle \nabla {\cal L}(\bm\theta) \delta \bm\theta^T \rangle = \eta^{-1} \text{tr}\, D_0,
\end{equation}
where we used $\bm\nabla {\cal L}(\bm\theta) \approx H_0 \delta \bm{\theta}$.  Eq.~\eqref{n3b} is equivalent to the first fluctuation-dissipation relation derived in Ref.~\citen{yaida2018fluctuation}. 

Eq.~\eqref{n3} can be solved for $\Sigma$, giving
\begin{equation}\label{n4}
\Sigma = O\Delta O^T.
\end{equation}
Here $O$ is an $N\times N$ matrix whose columns are eigenvectors of $H_0$, so that $(O^T H_0 O)_{\alpha\gamma} = h_\alpha \delta_{\alpha\beta}$, where $h_\alpha$ are the eigenvalues of $H_0$.  Since $\bm\theta_0$ is a local minimum, $h_\alpha >0$ for all $\alpha$.  $\Delta$ is another $N \times N$ matrix with elements $\Delta_{\alpha\beta} = 2\eta^{-1} (h_\alpha+h_\beta)^{-1} (O^T D^0 O)_{\alpha\beta}$.

With the above stationary solution, the velocity field of Eq.~\eqref{fp2b} becomes
\begin{equation}
\label{n5}
    \bm{v}^s(\bm\theta) = -C \Sigma^{-1} \delta\bm\theta, \quad \text{where} \quad C \equiv \eta H_0 \Sigma -D_0.
\end{equation}
Using Eq.~\eqref{n3} one can equivalently express $C$ as
\begin{equation}
    \label{n6}
    C = \frac{1}{2}\eta(H_0 \Sigma - \Sigma H_0),
\end{equation}
in which form it is apparent that $C$ is antisymmetric.  Following Ref.~\citen{tomita1974irreversible}, we call $C$ the circulation matrix, since it characterizes the stationary rotational flow of the probability around $\bm\theta_0$.  Equilibrium is only possible when $C = 0$, but this is generally not the case for SGD dynamics.  The circulation matrix is also related to the two trajectory quantities discussed in Sec.~\ref{trajpic}, the area matrix $A_{\alpha\gamma}(\Omega)$ and entropy production $\sigma(\Omega)$.  Imagine we have an SGD trajectory with $\ell$ steps, starting at some $\bm\theta_{t}$ drawn from the stationary distribution:  $\Omega = \{\bm\theta_t, \bm\theta_{t+1},\ldots,\bm\theta_{t+\ell}\}$.  The discrete analogue of Eq.~\eqref{tr2} is:
\begin{equation}
    \label{n7}
    \begin{split}
    A_{\alpha\gamma}(\Omega) &= \frac{1}{2}\sum_{k=t}^{t+\ell-1}  \bigl[\delta\theta_{k,\alpha} (\theta_{k+1,\gamma}-\theta_{k,\gamma})\\
    &\qquad\qquad- \delta\theta_{k,\gamma} (\theta_{k+1,\alpha}-\theta_{k,\alpha})\bigr],
    \end{split}
\end{equation}
where $\theta_{k,\alpha}$ denotes the $\alpha$th component of $\bm\theta_k$.  Similarly the discrete version of Eq.~\eqref{tr3} takes the form:
\begin{equation}
    \label{n8}
    \sigma(\Omega) = \sum_{k=t}^{t+\ell-1} \bm{v}^s(\bm\theta_k) \cdot D_0^{-1} (\bm\theta_{k+1}-\bm\theta_{k}).
\end{equation}
We can imagine averaging both quantities over an ensemble of stationary SGD trajectories, each of the same length $\ell$.  The mean area production per step turns out to be equal to the corresponding component of $C$~\cite{ghanta2017fluctuation}:
\begin{equation}
    \label{n9}
    \frac{\langle A_{\alpha\gamma} \rangle}{\ell}  = C_{\alpha\gamma},
\end{equation}
justifying the interpretation of $C$ as a circulation matrix.  In fact, we can interpret the magnitude of $C_{\alpha\gamma}$ as the angular momentum for the probability current projected onto the $\theta_\alpha$, $\theta_\gamma$ plane.  $C$ also plays a role in the mean entropy production per step~\cite{tomita2008irreversible},
\begin{equation}
    \label{n10}
    \frac{\langle \sigma \rangle}{\ell}  = -\text{tr}(C D_0^{-1} C \Sigma^{-1}).
\end{equation}

\section{Results and Discussion}\label{results}

\begin{figure}[ht!]
    \centering
    \includegraphics[width=0.93\columnwidth]{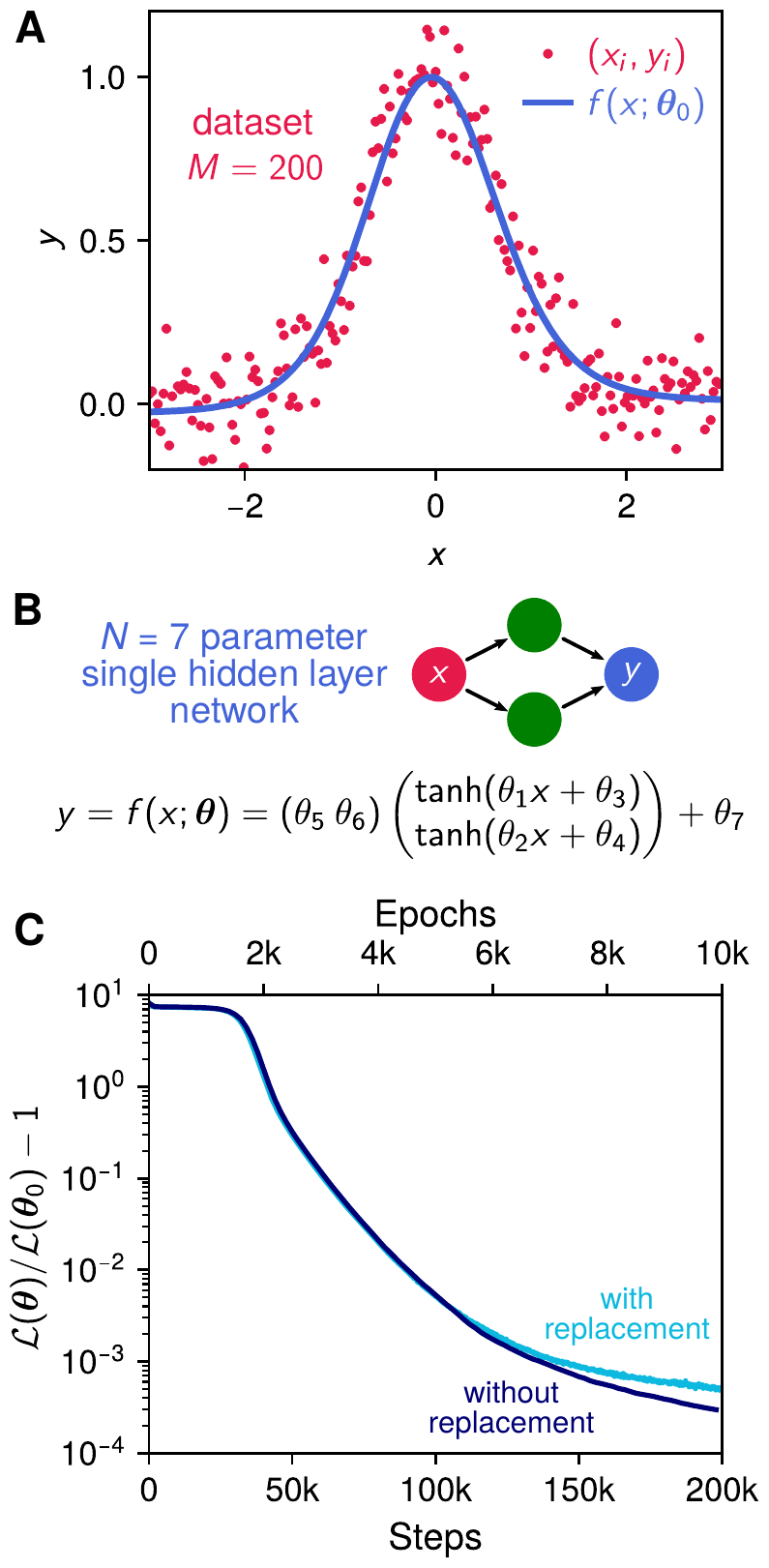}
    \caption{A) The dataset $(x_i, y_i)$, $i=1,\ldots,M$ (red points) for the simple nonlinear regression network example, and the network output $f(x;\bm\theta_0)$ (blue curve) for the parameter set $\bm\theta_0$ that minimizes the loss function. B) The network structure and functional form $f(x;\bm\theta)$, consisting of a single hidden layer of two neurons and tanh activation. C) Training results showing relative error of the loss ${\cal L}(\bm\theta)$ compared to the minimum ${\cal L}(\bm\theta_0)$, averaged over 1000 independent training runs.  With (WR) and without (WOR) replacement cases are shown as light and dark blue respectively.  The learning rate $\eta = 10^{-7}$, and a minibatch size $m=10$ was used for the WOR case.}
    \label{gaussian_model}
\end{figure}

\subsection{Nonlinear regression network}\label{sec:nonl_regnet}

To illustrate the theory, we mainly focus on a particularly simple neural network designed to solve a nonlinear regression task.  This example captures all the essential qualitative features of the theory, but the formalism is completely general and works for any network function $\bm{f}(\bm{x}_i;\bm\theta)$.  (As described below, we also show additional results for a more complex network designed to classify MNIST digits.)  The input data points in the simple example are given by $M=200$ scalar values $x_i = -3 + 0.03(i-1)$, $i=1,\ldots,M$.  The output values are given by $y_i = \exp(-x_i^2) + \eta_i$, where the $\eta_i$ are drawn from a Gaussian distribution of mean zero and standard deviation $\epsilon = 0.1$.  The $z_i = (x_i,y_i)$ dataset is illustrated in Fig.~\ref{gaussian_model}A.  As shown in Fig.~\ref{gaussian_model}B, the network function $f(x;\bm\theta)$ consists of a single hidden layer with two neurons, tanh activation, and $N=7$ parameters $\theta_\alpha$:
\begin{equation}\label{mod1}
f(x;\bm\theta) = \begin{pmatrix} \theta_5 & \theta_6 \end{pmatrix} \begin{pmatrix} \tanh(\theta_1 x + \theta_3) \\ \tanh(\theta_2 x + \theta_4) \end{pmatrix} + \theta_7.
\end{equation}
The loss function takes the form of Eq.~\eqref{sgd1} with mean squared error terms $\ell(z_i; \bm\theta) = \frac{M}{2\epsilon^2} (y_i - f(x_i;\bm\theta))^2$ and an L$_2$ regularization term $R(\bm\theta) = \lambda \bm\theta^2$, with $\lambda = 10$.  The loss function has a unique minimum, which we denote $\bm\theta_0$.  The network output at this minimum is shown as a blue curve in Fig.~\ref{gaussian_model}A, and amounts to a nonlinear regression fit to the noisy dataset.

The network parameters $\bm\theta$ can be trained via SGD, either using WR or WOR minibatching.  We choose a learning rate $\eta = 10^{-7}$, and a minibatch size $m = 10$, which defines an epoch as having $n = M/m = 20$ training steps.  The quantity ${\cal L}(\bm\theta)/{\cal L}( \bm\theta_0) -1$, the relative error with respect to the minimum loss value ${\cal L}(\bm\theta_0)$, is plotted as a function of training steps in Fig.~\ref{gaussian_model}C.  The two curves are the results for WR and WOR, in each case averaged over 1000 independent training runs of $2\times 10^5$ steps.  After an initial plateau, both kinds of training converge rapidly toward the vicinity of the minimum loss, with WOR slightly outperforming WR near the end of training.  The reason for this is connected to the magnitude of the fluctuations in the WOR versus WR stationary distribution, as we will see in Sec.~\ref{ness_res}.

\subsection{MNIST classification network}

As additional validation of the theory, we also trained a linear network to classify the MNIST database of handwritten digits~\cite{deng2012mnist}.  The training set is $M=60,000$ grayscale images, coarse-grained from their original size of 28$\times$28 pixels to 7$\times$7.  Thus the input data points $\bm{x}_i$ are vectors of dimension 49 (the flattened images), $\bm{f}(\bm{x}_i;\theta) = W(\bm\theta) \bm{x}_i$, and $W(\bm\theta)$ is a $10\times 49$ weight matrix whose elements comprise the $N = 490$ parameter vector $\bm\theta$.  The loss function consists of terms $\ell(z_i;\bm\theta) = (\bm{y}_i - \bm{f}(\bm{x}_i;\bm\theta)^2$, where $\bm{y}_i$ is a ten-dimensional vector corresponding to the one-hot encoded label of the digit (0-9).  As in the regression model, we use an L$_2$ regularization term, with $\lambda = 10^{-2}$.  For the results shown later, the learning rate $\eta = 10^{-4}$ and the minibatch size $m=100$, so there are $M/m = 600$ training steps per epoch.

\begin{figure*}[ht!]
    \centering
    \includegraphics[width=\textwidth]{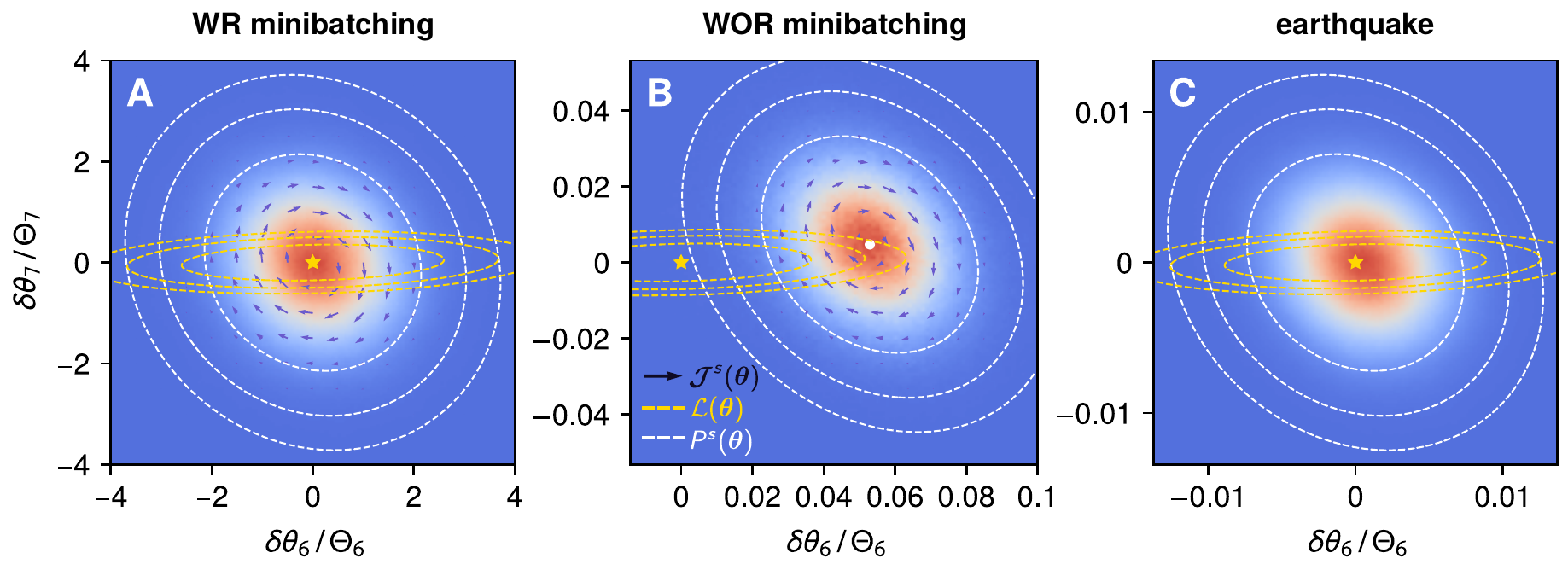}
    \caption{Cross-sections of the stationary distribution of network parameters for the regression example, under three different training approaches: A) with replacement (WR) minibatching; B) without replacement (WOR) minibatching; C) the earthquake model (Sec.~\ref{quake}).  In all cases we plot the marginal distribution in terms of the $\alpha = 6,7$ components of $\delta\bm\theta = \bm\theta - \bm\theta_0$, where $\bm\theta_0$ is the minimum of ${\cal L}(\bm\theta)$.   This point is the origin of each panel, and is marked by a yellow star.  The scale is set by $\bm\Theta_\alpha = \Sigma_{\alpha\alpha}^{1/2}$, the theoretical covariance matrix elements for the WR case, and is the same in all the panels to facilitate comparison.  The heat map (red = high, blue = low) shows the numerical distribution derived from training, while the white dashed contours are the equiprobability lines of $P^s(\bm\theta)$ derived from the theory (with a factor of 10 difference between each line).  For contrast, the contours of a Boltzmann distribution, which correspond to levels of the loss function ${\cal L}(\bm\theta)$, are shown as yellow dashed lines.  In the WR and WOR cases, which are out of equilibrium, the nonzero stationary current $\bm{\mathcal J}(\bm\theta)$ is depicted as a vector field.  The theoretical prediction for the center of the WOR distribution is shown as a white dot in panel B.}
    \label{dist}
\end{figure*}

\subsection{Nonequilibrium stationary state distributions}\label{ness_res}

As SGD training progresses, the distribution of network parameters converges to a stationary state, allowing us to numerically check our theoretical predictions from Sec.~\ref{ness}.  This NESS distribution $P^s(\bm\theta)$ is illustrated as a heat map in Fig.~\ref{dist}A,B for the WR and WOR cases respectively of the regression network.  The figures show a two-parameter projection of the full seven-dimensional distribution (i.e. marginalizing over the remaining five parameters), with axes $\delta \theta_6 = \theta_6 - \theta_{06}$, $\delta \theta_7 = \theta_7 -\theta_{07}$, where $\theta_{0i}$ is the $i$th component of $\bm\theta_0$.  Though we will describe the results for this particular projection, the qualitative details are similar if the distribution was plotted for any choice of $\delta \theta_i$ vs. $\delta \theta_j$.

To allow direct comparison of the fluctuations between the WR and WOR cases, we have the same units in all the panels of Fig.~\ref{ness}, using $\Theta_6 \equiv \Sigma_{66}^{1/2}$ and $\Theta_7 
 \equiv \Sigma_{77}^{1/2}$, components of the WR covariance matrix, as reference values for the scale.    Superimposed on the heat maps of $P^s(\bm\theta)$ are white dashed curves showing contours of equal stationary probability, with each successive contour corresponding to a factor of 10 decrease in probability.  In contrast, the expectation for a Boltzmann equilibrium distribution, $P^s(\bm\theta) \propto \exp(-\beta {\cal L}(\bm\theta))$, where the probability contours would have equal ${\cal L}(\bm\theta)$ values, are shown as gold dashed curves.  The clear discrepancy between the stationary distribution and the equilibrium expectation is consistent with the nonequilibrium nature of the stationary state (though as we will see below in Sec.~\ref{quake} there is more to the story).  The absence of equilibrium is directly reflected in the nonzero values of the stationary current density $\bm{\mathcal J}^s(\bm{\theta})$, indicated by the arrows in Fig.~\ref{dist}A,B, showing a net circulation.  

There are two main differences apparent in the WR and WOR cases.  The first is that the minimum $\bm\theta_0$ of the loss function ${\cal L}(\bm\theta)$, chosen as the origin in Figs.~\ref{ness}A and B (marked by a yellow star) corresponds to the center of the distribution for WR, but not for WOR.   Because the WOR fluctuations correspond to the effective loss landscape $\hat{\cal L}(\bm\theta)$ from Eq.~\eqref{sgd4}, they are centered around a new minimum $\hat{\bm\theta}_0$.  The $\hat{\bm\theta}_0$ calculated from minimizing Eq.~\eqref{sgd4} is shown as a white circle in Fig.~\ref{dist}B, and agrees well with the center of the distribution derived from simulations.  Because the correction term that distinguishes $\hat{\cal L}(\bm\theta)$ from ${\cal L}(\bm\theta)$ is of order ${\cal O}(\eta)$, the shift in the minima $\hat{\bm\theta}_0 - {\bm\theta}_0$ does not significantly impact the ability of the system to learn the ``true'' solution ${\bm\theta}_0$ determined by the original loss function.  In fact, as seen in Fig.~\ref{gaussian_model}C, the WOR training on average finds network weights $\bm\theta$ that are {\it closer} to $\bm\theta_0$ than the WR case.

\begin{figure}[ht!]
    \centering
    \includegraphics[width=\columnwidth]{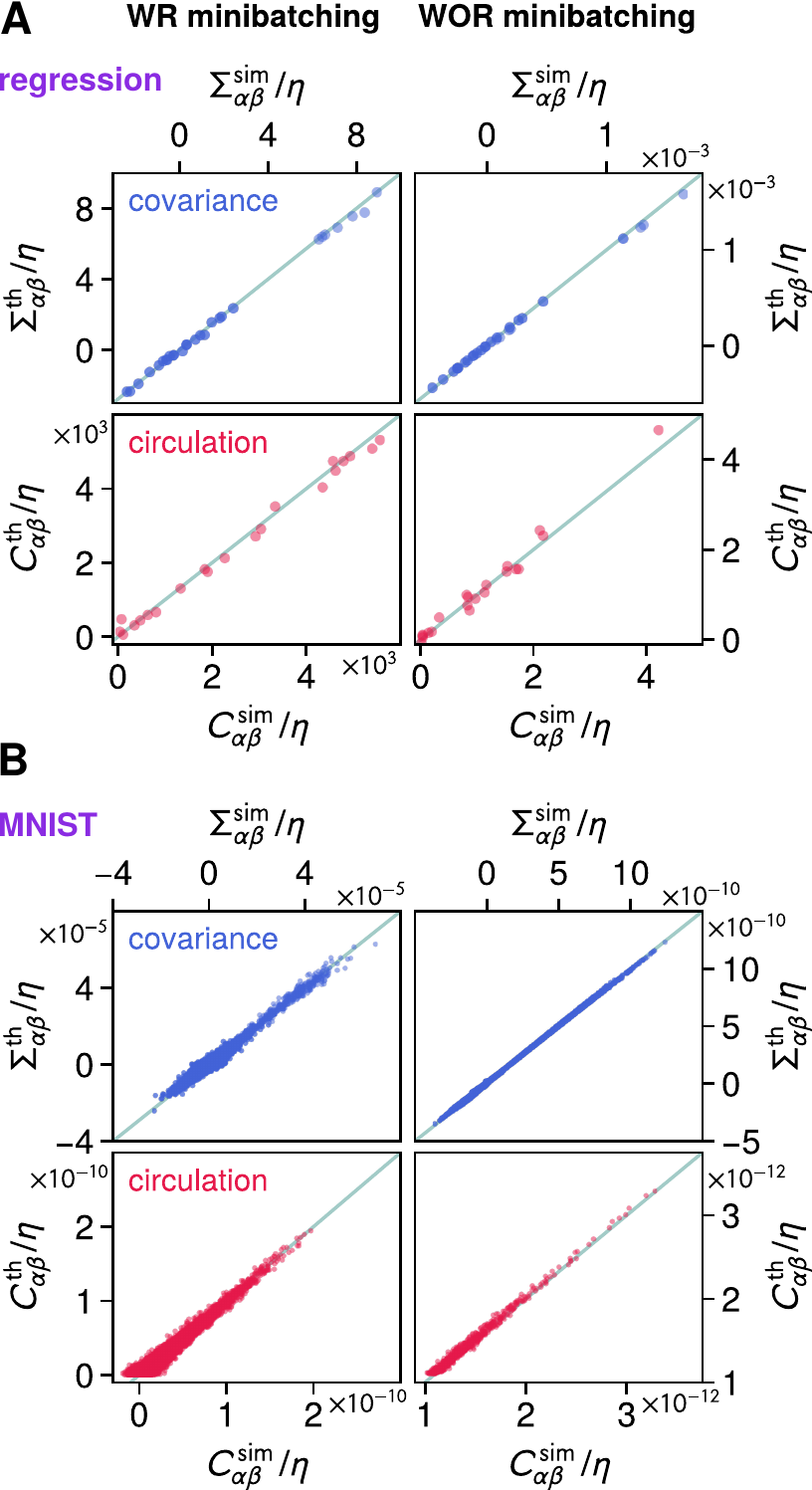}
    \caption{Elements of two matrices characterizing the nonequilibrium stationary state for the A) nonlinear regression network, and B) MNIST.  In each case we show the covariance matrix elements $\Sigma_{\alpha\beta}$, $\alpha,\beta=1,\ldots,7$ (blue) and the circulation matrix elements $C_{\alpha\beta}$ (red) for WR (left column) and WOR (right column) minibatching.  The horizontal axes (``sim'' superscript) indicate values derived from simulations of $\sim 10^7 - 10^8$ steps. The vertical axes (``th'' superscripts) indicate theoretical predictions:  Eqs.~\eqref{n4} and \eqref{n6} for $\Sigma$ and $C$ respectively.  The diagonal line corresponds to perfect agreement between theory and simulations.  Note the difference in magnitude scales between the WR and WOR quantities.}
    \label{checks}
\end{figure}

The explanation for this seemingly strange result is that the fluctuations for WOR are two order of magnitudes smaller in this example compared to WR, as can be seen by comparing the scales on the axes of Fig.~\ref{dist}A and B.  We can also see this explicitly in the elements of the $7 \times 7$ covariance matrix $\Sigma_{\alpha\beta}$.  The top panels of Fig.~\ref{checks}A show all $N(N+1)/2 = 28$ distinct elements for the WR and WOR cases respectively, comparing the simulation-derived values versus the theoretical prediction of Eq.~\eqref{n4}.  Simulation and theory show excellent agreement.  The $\Sigma_{\alpha\beta}$ elements in the WR and WOR cases differ by almost four order of magnitudes across the whole 7-dimensional parameter space, corresponding to about two orders of magnitude difference in root-mean-square fluctuations, consistent with the results in Fig.~\ref{dist}A,B.

The same qualitative story holds for the MNIST network.  The top panels of Fig.~\ref{checks}B show the $N(N+1)/2 = 120,295$ distinct elements of $\Sigma_{\alpha\beta}$ for the WR and WOR cases, and again the agreement between simulation and theory is very good.  Just as in the regression example, the covariance scale is much smaller in the WOR case, by about four orders of magnitude.

\begin{figure}[ht!]
    \centering
    \includegraphics[width=\columnwidth]{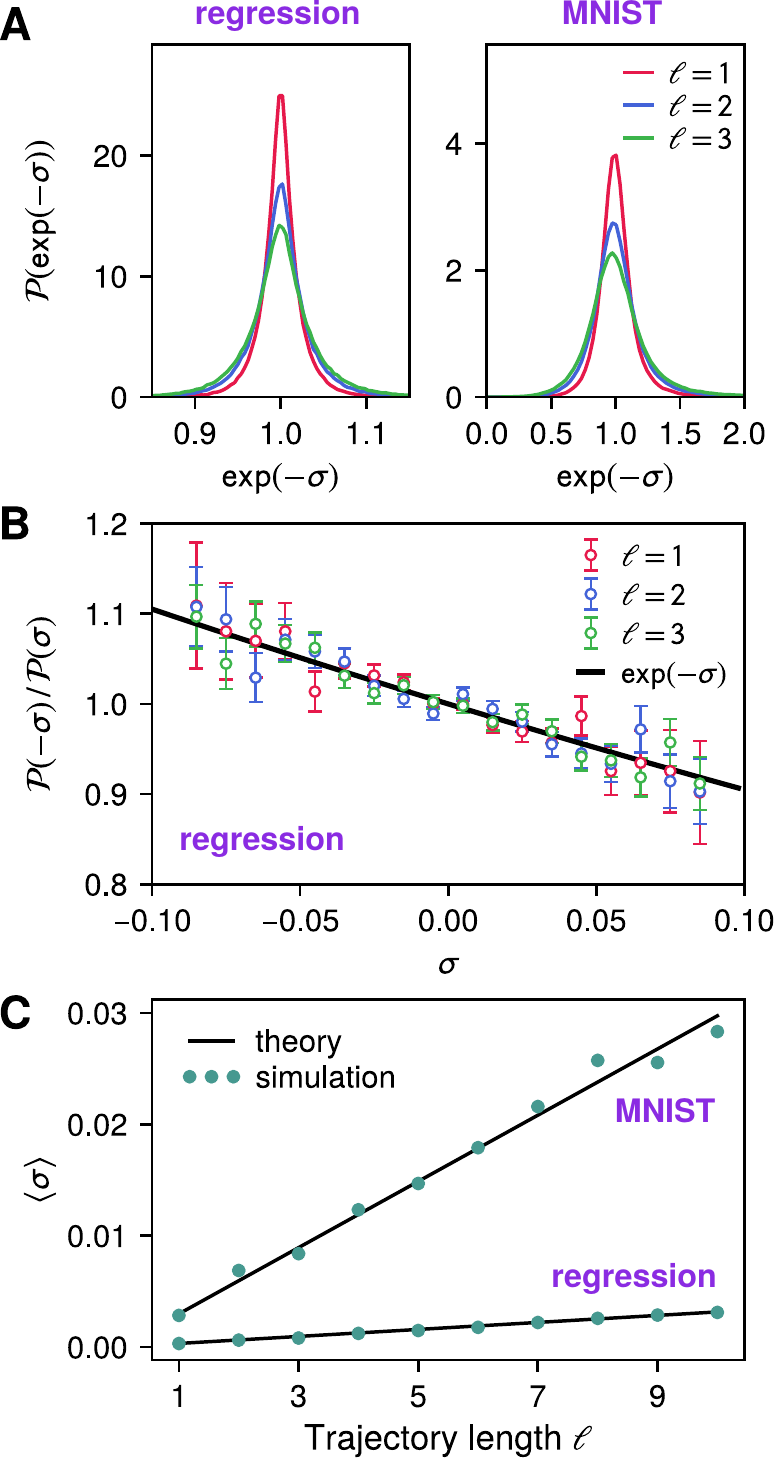}
    \caption{Properties of entropy production $\sigma$ in the stationary state.  A) The distribution of $\exp(-\sigma)$ for numerical trajectories of length $\ell =1-3$ for the regression (left) and MNIST (right) networks.  Consistent with the IFT, these distributions have averages that are approximately equal to 1.  B) Relative likelihood ${\cal P}(-\sigma)/{\cal P}(\sigma)$ versus $\sigma$ for regression with $\ell =1-3$.  The DFT predicts this ratio to be $\exp(-\sigma)$, shown as a black curve. (C) Mean $\langle \sigma \rangle$ as a function of $\ell$ for the MNIST (top) and regression (bottom) cases.  The corresponding theoretical predictions of Eq.~\eqref{n10} are shown as solid lines.  For both networks we show results using WR minibatching, but the WOR case would be qualitatively similar.}
    \label{ift}
\end{figure}

\subsection{Circulation and fluctuation theorems}

Two final aspects of the NESS distribution are interesting to highlight:  the net circulation and the integral fluctuation theorem (IFT), both reflecting the breaking of a fundamental symmetry---the likelihoods of observing a sequence of states and the reverse sequence no longer being equal, as is the case in equilibrium.  The circulation matrix $C$ is theoretically defined in Eq.~\eqref{n6}, but can also be estimated from simulation trajectories using the area matrix of Eq.~\eqref{n7}, whose trajectory average can be related to the $C$ via Eq.~\eqref{n9}.  The bottom panels of Fig.~\ref{checks}A,B compare theory versus simulation results for the elements of $C$ in the regression and MNIST networks, showing close agreement.  As with $\Sigma_{\alpha\beta}$, there are significant differences in the order of magnitude of $C_{\alpha\beta}$ between the WR and WOR cases in both examples.

The nonzero circulation $C$ is related, via Eq.~\eqref{n10}, to nonzero mean entropy production $\sigma(\Omega)$.  If we divide up the long simulation trajectory into subtrajectories $\Omega$ of length $\ell$, and calculate $\sigma(\Omega)$ for each subtrajectory using Eq.~\eqref{n8}, the mean $\langle \sigma \rangle$ over the whole ensemble will always be greater than zero for an NESS (in accord with the ``second law'' analogy).  We show the simulation results for $\langle \sigma \rangle$ versus $\ell$ as points in Fig.~\ref{ift}C, validating the theoretical prediction of Eq.~\eqref{n10} (solid lines, with MNIST at the top and regression at the bottom, in both cases using WR minibatching).  The fact that $\langle \sigma \rangle > 0$ means that the probability distribution ${\cal P}(\sigma)$ must be asymmetric, with more weight in subtrajectories that produce entropy rather than destroy it.  Entropy-producing $\Omega$ have $\sigma(\Omega) >0$, corresponding to a subtrajectory mostly aligned with the circulation direction, while in the entropy-destroying case $\sigma(\Omega) <0$, corresponding to an $\Omega$ mostly anti-aligned to the circulation.  However entropy production in the NESS also has less apparent properties:  according to the IFT the distribution of exponentiated entropies, ${\cal P}(e^{-\sigma})$, must have an average of 1.  ${\cal P}(e^{-\sigma})$ distributions for three different values of $\ell$ are shown in Fig.~\ref{ift}A for the regression (left) and MNIST (right) examples.  The numerically calculated average $\langle e^{-\sigma} \rangle$ for the various $\ell$ shown is within $0.02\%$ of 1 for the regression case, and within $2\%$ of 1 for the MNIST case.  The IFT is in turn closely related to the DFT, Eq.~\eqref{tr5}, which dictates the relative likelihood ${\cal P}(-\sigma)/{\cal P}(\sigma) = \exp(-\sigma)$.  Fig.~\ref{ift}B shows this ratio as a function of $\sigma$ for various $\ell$ in the regression case, and it is consistent (within error bars) with the DFT expectation of $\exp(-\sigma)$.  Overall, we thus numerically verify the thermodynamic analogy described in Sec.~\ref{ness}: given the definition of entropy production in Eq.~\eqref{n8}, the network weight fluctuations induced by SGD have the essential characteristics of a physical nonequilibrium stationary state.

\subsection{Inverse variance-flatness relation and the earthquake model}\label{quake}

The discrepancy between the equiprobability contours of the stationary distribution $P^s(\bm\theta)$ (white dashed curves) and the contours of ${\cal L}(\bm\theta)$ (gold dashed curves), seen in Fig.~\ref{dist}A,B, has been observed before in SGD dynamics. Feng and Tu dubbed it the ``inverse variance-flatness''~\cite{feng2021inverse} relation:  contrary to the equilibrium expectation, there is a larger variance in the weight fluctuations along directions where the loss landscape ${\cal L}(\bm\theta)$ is {\it less} flat.  What drives this counterintuitive preference for climbing up steep walls?  At a qualitative level, consider a trajectory traversing a narrow valley in the loss landscape surrounded by steep sides.  The combination of a finite step size and the fact that minibatching gives an approximation to the true gradient means that rather than flowing smoothly along the valley bottom, the trajectory will often zig-zag between the steep walls.  A strong repulsive gradient force at one wall will tend to make it overshoot the valley and end up hitting another steep wall.  The end result is a distribution that preferentially samples directions with steep sides.  In the case of conventional SGD (WR or WOR minibatching) this inverse-flatness relation is associated with the presence of nonzero currents.  But interestingly, we can find similar behavior in an idealized equilibrium model of training dynamics, where the net current is zero.  Here we introduce this alternative ``earthquake'' model as a useful comparison to conventional SGD, replicating some of the qualitative features of the SGD stationary state while remaining in equilibrium.

Training in the earthquake model proceeds through a modified gradient descent equation:
\begin{equation}
\label{quake1}
\bm{\theta}_{t+1} = \bm{\theta}_{t} - \eta\,  \bm{\nabla} {\cal L}(\bm{\theta}_t + \zeta \bm z_t).
\end{equation}
Here $\bm z_t$ is a random $N$-dimensional Gaussian vector with $\langle \bm z_t \rangle = 0$, $\langle \bm z_t \bm z_t^T \rangle = I$ and $\zeta$ is the noise magnitude.  The addition of a random displacement to the argument of the loss function means the gradient is calculated from a loss landscape that is shifted in a random direction at each time step (hence the ``earthquake'' moniker).  Unlike the random landscape theory described in Ref.~\citen{feng2021inverse}, the shape of the landscape is preserved at each step, which makes the resulting dynamics easy to analyze.

Let us focus on stationary state fluctuations around $\bm\theta_0$ in the limit of small noise $\zeta$.  In this regime the gradient can be approximated via $\bm\nabla {\cal L}(\bm\theta_t + \zeta \bm z_t) \approx H_0( \delta \bm{\theta} + \zeta \bm z_t)$.  Eq.~\eqref{quake1} can then be rewritten as
\begin{equation}
\label{quake2}
\bm{\theta}_{t+1} \approx \bm{\theta}_{t} - \eta H_0 \delta \bm{\theta}_t + \bm\xi_t,
\end{equation}
where the stochastic contribution $\bm\xi_t \equiv - \eta \zeta H_0 \bm z_t$. By inserting an identity matrix $I = H_0 H_0^{-1}$ before the $\delta \bm\theta_t$, we can recast Eq.~\eqref{quake2} as equilibrium dynamics on an effective landscape ${\cal L}^\text{quake}(\bm\theta)$ whose Hessian is the inverse of the original $H_0$,
\begin{equation}
\label{quake3}
\begin{split}
\bm{\theta}_{t+1} &\approx \bm{\theta}_{t} - \eta H_0^2 H_0^{-1}\delta \bm{\theta}_t + \bm\xi_t\\
&= \bm{\theta}_{t} - \mu \bm\nabla {\cal L}^\text{quake}(\bm{\theta}_t) + \bm\xi_t.
\end{split}
\end{equation}
where
\begin{equation}\label{lquake}
    {\cal L}^\text{quake}(\bm{\theta}_t) = {\cal L}(\bm\theta_0) + \frac{1}{2}\delta \bm\theta^T H_0^{-1} \delta\bm\theta.
\end{equation}
The corresponding mobility matrix $\mu = \eta H_0^2$ and diffusion matrix $D = \frac{1}{2}\langle \bm \xi_t \bm \xi_t^T\rangle = \frac{1}{2} \eta^2 \zeta^2 H_0^2$ now satisfy the Einstein relation $\mu = \beta D$ with $\beta = \eta \zeta^2/2$.  Hence the stationary state for the earthquake model is an equilibrium Boltzmann distribution, but on an effective landscape: $P^s(\bm\theta) \propto \exp(-\beta {\cal L}^\text{quake}(\bm\theta))$.

Crucially, however, the Hessian of 
${\cal L}^\text{quake}(\bm\theta)$ in Eq.~\eqref{lquake}
is the inverse of the Hessian $H_0$ of the original landscape.  This will directly lead to an inverse variance-flatness relation, qualitatively quite similar to that observed for SGD dynamics.  We can see this by training the regression network using the earthquake approach (with $\zeta = 10^{-4}$), yielding the stationary distribution seen in Fig.~\ref{dist}C.  Despite the absence of any nonequilibrium currents, fluctuations are extended in directions corresponding to steep gradients in the original landscape, and suppressed in shallow directions.  Of course, by construction, the stochasticity in the earthquake model is much simpler than for SGD minibatching.  However the underlying mechanism is similar: a finite step size combined with perturbations $\bm{z}_t$ means that the trajectory can wander onto a steep slope, where it will likely get a significant push back, overshooting the valley and ending up on another steep slope.  Here the particular structure of the noise induced by training leads to an equilibrium on an effective landscape significantly different from the original.  We saw an earlier example of a noise-induced effective landscape $\hat{\cal L}(\bm{\theta})$ in the WOR minibatching case, though with a nonequilibrium stationary state.  Taking into account the shape of these noise-induced landscapes can have practical ramifications, as we demonstrate in the next section, detailing a Bayesian machine learning application.

\subsection{Exploiting noise during training:  SGWORLD algorithm for improved Bayesian learning via WOR minibatching}\label{sec:sgworld}

In certain applications the loss function has a Bayesian interpretation, with the two parts ${\cal L}(\bm\theta) = L(\bm\theta) + R(\bm\theta)$ in Eq.~\eqref{sgd1} related to the likelihood ${\cal P}_\text{lk}({\cal D} | \bm\theta )$ of the training data ${\cal D} = \{ (\bm{x}_i,\bm{y}_i), i=1,\ldots,M\}$ given the network parameters $\bm\theta$, and the prior distribution ${\cal P}_\text{pr}(\bm\theta)$ of the parameters:
\begin{equation}
    \label{bay1}
    {\cal P}_\text{lk}( {\cal D} | \bm\theta ) = e^{-L(\bm\theta)}, \quad {\cal P}_\text{pr}(\bm\theta) = e^{-R(\bm\theta)}.
\end{equation}
From Bayes' rule the corresponding posterior distribution is
\begin{equation}
    \label{bay2}
    {\cal P}_\text{po}( \bm\theta | {\cal D}) \propto {\cal P}_\text{lk}( {\cal D} | \bm\theta )  {\cal P}_\text{pr}(\bm\theta) = e^{-{\cal L}(\bm\theta)},
\end{equation}
up to a normalization constant that ensures $\int d\bm\theta\, {\cal P}_\text{po}( \bm\theta | {\cal D}) = 1$.  In this Bayesian framework a network $\bm\theta_0$ that minimizes ${\cal L}(\bm\theta)$ also maximizes the posterior.  But we may be interested not just in a single network, but in an ensemble of networks drawn from ${\cal P}_\text{po}( \bm\theta | {\cal D})$.  Constructing such an ensemble allows us to for example to quantify confidence in network predictions, a key goal of Bayesian machine learning~\cite{ghahramani2015probabilistic}.

To create this ensemble, we imagine a two-stage training process:  in the first stage, we train using a conventional method (such as SGD) that brings us to the vicinity of $\bm\theta_0$.  In the second stage we modify the training procedure to ensure the stationary distribution $P^s(\bm\theta) \approx {\cal P}_\text{po}( \bm\theta | {\cal D})$, so that training steps approximate as closely as possible draws from the posterior.  Can we use our knowledge of how training noise influences $P^s(\bm\theta)$ to engineer an effective protocol for this second stage?

Perhaps the most straightforward approach to achieve this is based on stochastic gradient Langevin dynamics (SGLD)~\cite{welling2011bayesian}, whose algorithm is summarized in the left column of Fig.~\ref{alg}.  We run the second-stage training for $N_\text{epoch}$ epochs, with each epoch consisting of $n$ steps.  At each step we draw batches $B_t$ with replacement, and perform ordinary WR SGD as described in Sec.~\ref{sec:sgd_wr}.  The one modification is that we add noise $\sqrt{2\eta} \bm{z}_t$ at the end of each step, where $\bm{z}_t$ is an $N$-dimensional vector of Gaussian noise with unit variance.  Carrying out the analogous derivation of Sec.~\ref{sec:sgd_wr}, we can map the system onto a Fokker-Planck equation with modified mobility and diffusion matrices:
\begin{equation}
    \label{bay3}
    \mu'(\bm\theta) = \eta I, \qquad D'(\bm\theta) = \eta I + D(\bm\theta),
\end{equation}
where $D(\bm\theta)$ is given by Eq.~\eqref{sgd3}.  Since $D(\bm\theta) \sim {\cal O}(\eta^2)$, we see that $D(\bm\theta)$ becomes negligible relative to $\eta I$ in the limit $\eta \to 0$.  Hence in this limit SGLD dynamics approximately satisfies the Einstein relation with $\beta = 1$, meaning that $P^s(\bm\theta) \propto \exp(-{\cal L}(\bm\theta)) = {\cal P}_\text{po}(\bm\theta|{\cal D})$.  In fact the original SGLD algorithm included a scheme for progressively decreasing $\eta$ at each step to ensure this limit is satisfied (with the potential tradeoff of slower convergence to the stationary state).  Here we will employ the often-used simplification of keeping $\eta$ fixed (at a small value $\eta \ll 1$) during training.  However in the example described below we will study the accuracy of the algorithm at different magnitudes of $\eta$.

\begin{figure*}[t]
\centering
\begin{subfigure}[t]{0.45\textwidth}
\vskip 0pt
\hrule\vspace{0.25em}
{\bf SGLD}: stochastic gradient Langevin dynamics~\cite{welling2011bayesian}
\vspace{0.25em}
\hrule
\vspace{0.25em}
\begin{algorithmic}
\State $t \gets 0$
\For{$j=1,\ldots,n N_{\text{epoch}}$}
\State Choose batch $B_t$ with replacement.
\LComment{SGD step:}
\State $\bm\theta_{t+1} \gets \bm\theta_t - \eta \bm\nabla L_{B_t}(\bm\theta_t) - \eta \bm\nabla R(\bm\theta_t)$
\State $t \gets t+1$
\LComment{$\bm{z}_t$ is Gaussian noise with unit variance:}
\State $\bm\theta_{t} \gets \bm\theta_{t} + \sqrt{2 \eta}\bm{z}_t$
\EndFor
\end{algorithmic}
\end{subfigure}
\hspace{0.02\textwidth}
\begin{subfigure}[t]{0.45\textwidth}
\vskip 0pt
\hrule\vspace{0.25em}
{\bf SGWORLD}: SGLD with WOR minibatching
\vspace{0.25em}
\hrule
\vspace{0.25em}
\begin{algorithmic}
\State $t \gets 0$
\For{$\tau=1,\ldots,N_{\text{epoch}}$}
\For{$j = 1,\ldots,n$}
\State Choose batch $B_t$ without replacement.
\LComment{SGD step:}
\State $\bm\theta_{t + 1} \gets \bm\theta_{t} - \eta \bm\nabla L_{B_t}(\bm\theta_{t}) - \eta \bm\nabla R(\bm\theta_{t})$
\State $t \gets t+1$
\EndFor
\LComment{$\bm{z}_t$ is Gaussian noise with unit variance:}
\State $\bm\theta_{t} \gets \bm\theta_{t} + \sqrt{2 \eta n}\bm{z}_t$
\LComment{Correction for effective WOR landscape:}
\State $\bm\theta_t \gets \bm\theta_t + \eta \bm\nabla \delta\!{\cal L}(\bm\theta_{t-n})$

\EndFor
\end{algorithmic}
\end{subfigure}
\caption{Two algorithms for sampling from the posterior distribution of network weights: SGLD~\cite{welling2011bayesian} and the proposed SGWORLD variant (details in the text).}
\label{alg}
\end{figure*}

The discussion of WOR SGD in Sec.~\ref{sec:sgd_wor} suggests that there may be an alternative, more accurate variant of SGLD based on WOR minibatching, which we will call SGWORLD.  We summarize this new algorithm in the right column of Fig.~\ref{alg}.  For each epoch $\tau = 1,\ldots,N_\text{epoch}$, we run $n$ SGD steps without replacement.  We add noise $\sqrt{2 n \eta} \bm{z}_t$ at the end of the epoch (equivalent to the sum of $n$ noise additions at each time step).  The only conceptual difference from the WR case comes from the fact that WOR dynamics epoch-to-epoch sees an effective landscape $\hat{\cal L}(\bm{\theta}) = n {\cal L}(\bm\theta) + \delta\!{\cal L}(\bm\theta)$ shown in Eq.~\eqref{sgd4}.  Since we want to draw from the posterior described by ${\cal L}(\bm\theta)$ rather than $\hat{\cal L}(\bm\theta)$, we need to correct for the gradient force that arises from the perturbation $\delta\!{\cal L}(\bm\theta)$.  Thus the last step of the algorithm is to add a correction force $\eta\bm\nabla\delta\!{\cal L}(\bm\theta_{t-n})$ at the end of the epoch.

The mapping to the Fokker-Planck equation gives us mobility and diffusion matrices of the form
\begin{equation}
    \label{bay4}
    \mu''(\bm\theta) = \eta I, \qquad D''(\bm\theta) = \eta n I + \hat{D}(\bm\theta),
\end{equation}
where $\hat{D}(\bm\theta)$ is given by Eq.~\eqref{sgd7}.  Since $\hat{D}(\bm\theta) \sim {\cal O}(\eta^4)$, it becomes negligible with respect to $\eta I$ much faster in the $\eta \to 0$ limit compared to $D(\bm\theta) \sim {\cal O}(\eta^2)$ in the SGLD case.  Thus in principle SGWORLD should provide a closer approximation to drawing from the posterior distribution than SGLD.

To test this hypothesis, we consider an example where the posterior distribution is exactly known:  a linearized version of the regression network in Sec.~\ref{sec:nonl_regnet}.  In this version the network function $f(x;\bm\theta)$ has a simplified form involving $N=3$ parameters $\theta_\alpha$:
\begin{equation}
    \label{bay5}
    f(x;\bm\theta) = \begin{pmatrix} \theta_1 & \theta_2 & \theta_3 \end{pmatrix} \begin{pmatrix} \tanh\left(\frac{1}{2}-x\right) \\ \tanh\left(\frac{1}{2}+x\right) \\ 1 \end{pmatrix} \equiv \bm\theta^T \bm\psi(x),
\end{equation}
where $\bm\psi(x)$ is a $3$-dimensional vector function.  As before, the loss function is given by
\begin{equation}
    \label{bay6}
    L(\bm\theta) = \frac{1}{2\epsilon^2}\sum_{i=1}^M \left(y_i - f(x_i;\bm\theta)\right)^2 = \frac{1}{2\epsilon^2}(\bm{y} - \bm\theta^T \Psi(\bm{x}))^2,
\end{equation}
with $\epsilon=0.1$.  Here $\bm{x}$ and $\bm{y}$ are $M$-dimensional vectors with components $x_i$ and $y_i$ respectively, and $\Psi(\bm{x})$ is a $3 \times M$ dimensional matrix with components $\Psi_{\alpha i}(\bm{x}) = \psi_\alpha(x_i)$.  The regularization term $R(\bm\theta) = \lambda \bm\theta^2$ with $\lambda = 10$.

Bayesian linear regression theory gives us the exact posterior distribution for this problem:
\begin{equation}
    \label{bay7}
    {\cal P}_\text{po}( \bm\theta | {\cal D}) \propto \exp\left(\frac{1}{2} \delta \bm\theta^T \Sigma^{-1}_\text{po} \delta\bm\theta \right),
\end{equation}
where the covariance matrix $\Sigma_\text{po}$ can be calculated from
\begin{equation}
    \label{bay8}
    \Sigma^{-1}_\text{po} = \frac{1}{\epsilon^2} \Psi(\bm{x}) \Psi^T(\bm{x}) + 2 \lambda I = H_0.
\end{equation}
Here $I$ is the $3 \times 3$ identity matrix and $H_0$ is the Hessian of ${\cal L}(\bm\theta) = L(\bm\theta) + R(\bm\theta)$.  The posterior is maximized at a value $\bm\theta_0$ given by
\begin{equation}
    \label{bay9}
    \bm\theta_0 = \frac{1}{\epsilon^2} \Sigma_\text{po} \Psi(\bm{x}) \bm{y}.
\end{equation}

\noindent {\it Estimating SGLD accuracy:} To calculate the effectiveness of the SGLD algorithm in predicting the true posterior, we can use a generalization of the stationary state theory in Sec.~\ref{sec:ss} (full details in Appendix D).  In the small $\eta$ limit the stationary distribution $P^s(\bm\theta)$ generated by SGLD for this example is approximately Gaussian, $P^s(\bm\theta) \propto \exp\left(\frac{1}{2} \delta \bm\theta^T \Sigma^{-1} \delta\bm\theta \right)$ with the same $\bm\theta_0$ as Eq.~\eqref{bay9}, but with a different covariance matrix satisfying
\begin{equation}
    \label{bay10}
    H_0 \Sigma +\Sigma H_0 = \eta^{-1} (2 D'(\bm\theta_0) + {\cal H}_{D'}(\bm\theta_0) \Sigma).
\end{equation}
The above equation generalizes the Lyapunov relation of Eq.~\eqref{n3} to include a correction induced by the variation of $D'(\bm\theta)$ in the vicinity of $\bm\theta_0$.  This variation is encoded in ${\cal H}_{D'}(\bm\theta)$, a fourth-order ``Hessian'' tensor associated with ${D'}(\bm\theta)$.  Its contraction with $\Sigma$ evaluated at $\bm\theta_0$, denoted as ${\cal H}_{D'}(\bm\theta_0) \Sigma$, is a $3 \times 3$ matrix with components
\begin{equation}
    \label{bay11}
    ({\cal H}_{D'}(\bm\theta_0) \Sigma)_{\alpha\beta} \equiv \sum_{\mu=1}^3 \sum_{\gamma=1}^3 \partial_\mu \partial_\gamma D'_{\alpha\beta}(\bm\theta_0) \Sigma_{\mu\gamma}.
\end{equation}
  Eq.~\eqref{bay10} is a system of equations for the components of $\Sigma$ which can be solved numerically. 
  
  Once $\Sigma$ is known, we can use the Kullback-Leibler (KL) divergence $D_\text{KL}(P^s || {\cal P}_\text{po} )$ to measure the difference between the true posterior distribution ${\cal P}_\text{po}(\bm\theta|{\cal D})$ and the SGLD distribution $P^s(\bm\theta)$:
\begin{equation}
\begin{split}
    \label{bay12}
    D_\text{KL}( P^s|| {\cal P}_\text{po}) &\equiv \int d\bm\theta\,P^s(\bm\theta) \log_2 \frac{P^s(\bm\theta)}{{\cal P}_\text{po}(\bm\theta|{\cal D})}\\
    &= \frac{1}{2\ln 2}\left[\ln \frac{\det \Sigma_\text{po}}{\det \Sigma} -3 + \text{tr}\,\left(\Sigma_\text{po}^{-1}\Sigma\right)\right].
\end{split}
\end{equation}
The KL divergence is a non-negative quantity (expressed in bits), and equals zero if the distributions are identical.

\begin{figure}[t]
\includegraphics[width=\columnwidth]{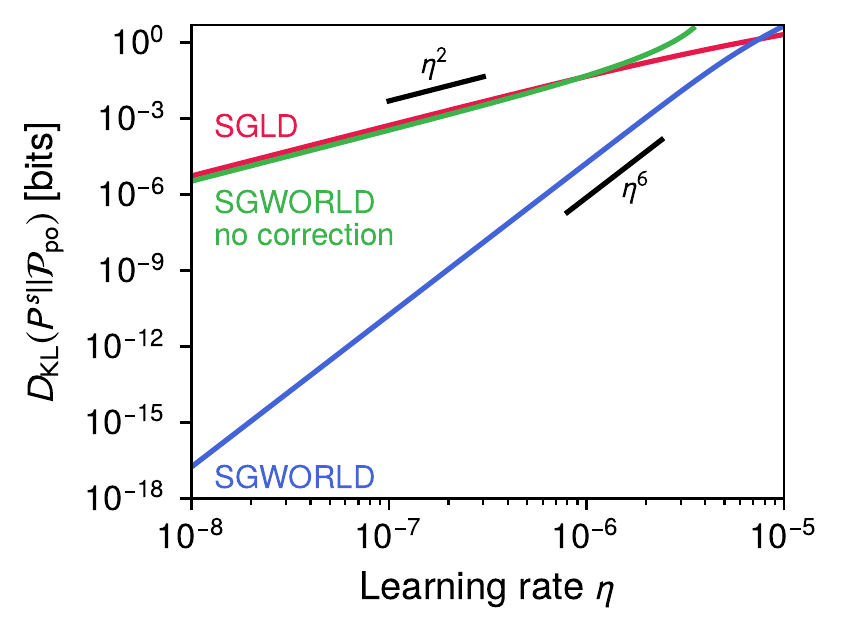}
\caption{The Kullback-Leibler (KL) divergence between the true posterior for the linearized regression network, and the posterior estimated via three different algorithms:  SGLD~\cite{welling2011bayesian} (red), SGWORLD (blue), and SGWORLD without the WOR landscape correction (green).  The divergence depends on the value of the learning rate $\eta$, and scales like a power law in $\eta$ for small $\eta$. 
 The SGWORLD algorithm shows the fastest convergence to the true posterior with decreasing $\eta$.}\label{bayes}
\end{figure}

Fig.~\ref{bayes} shows the KL divergence for SGLD (red curve) as a function of learning rate $\eta$.  The KL divergence scales like $\sim \eta^2$ for small $\eta$.  Though each value of the divergence here is calculated for a fixed $\eta$, this scaling would also govern the rate of convergence to the true posterior if $\eta$ was decreased dynamically during the training process.

\vspace{1em}

\noindent {\it Estimating SGWORLD accuracy:} The analogous calculation for the SGWORLD algorithm, finding the KL divergence between the stationary distribution and the true posterior, involves several modifications.  We use the WOR stationary state theory of Sec.~\ref{sec:sgd_wor}, but the additional correction force at the end of each of epoch corresponds to replacing $\hat{\cal L}(\bm{\theta})$ with a ``corrected'' effective landscape $\hat{\cal L}^c(\bm{\theta}) = n {\cal L}(\bm{\theta})$, which has the same minimum parameters $\bm\theta_0$ as ${\cal L}(\bm{\theta})$.

The SGWORLD stationary distribution takes the form $P^s(\bm\theta) \propto \exp\left(\frac{1}{2} \delta \bm\theta^T \hat\Sigma^{-1} \delta\bm\theta \right)$ with a covariance matrix $\hat\Sigma$ given by
\begin{equation}
    \label{bay13}
    n H_0 \hat\Sigma + n\hat\Sigma H_0 = \eta^{-1} (2 D''(\bm\theta_0) + {\cal H}_{D''}(\bm\theta_0) \hat\Sigma),
\end{equation}
 Finally we can calculate the KL divergence from Eq.~\eqref{bay12} with $\Sigma$ replaced by $\hat\Sigma$.

 The blue curve in Fig.~\ref{bayes} shows that the resulting KL divergence scales like $\eta^6$, and hence SGWORLD converges much more quickly than SGLD to the true posterior as $\eta$ is made smaller.  The effectiveness of SGWORLD is ultimately due to the reduced magnitude of minibatch-induced fluctuations for WOR versus WR.  However to take advantage of the suppressed fluctuations, one has to correct for another feature of WOR:  the effective perturbation to the loss landscape $\delta {\cal L}$.  Without the correction (the last line of the SGWORLD algorithm), the discrepancy induced by $\delta {\cal L}$ essentially cancels out the benefit of smaller fluctuations.  The green curve of Fig.~\ref{bayes} shows the results of the uncorrected SGWORLD algorithm, and they are only marginally better than SGLD in the limit of small $\eta$.  Note that both SGLD and SGWORLD algorithms require a sufficiently small $\eta$ to be numerically stable, and this threshold is problem-dependent.  For the regression example shown here, all the algorithms become unreliable at $\eta$ values $\gg 10^{-5}$, the right edge of Fig.~\ref{bayes} where the KL divergences become substantial.

\section{Conclusions}

Our analysis of SGD training of neural networks has revealed its close analogies to nonequilibrium physical systems---a stationary state that violates the Einstein relation, and fluctuations which obey universal and detailed fluctutation theorems.  At the same time, the characteristics of this stationary state are highly sensitive to technical aspects of the algorithm that at first glance might seem trivial.  Switching from with to without replacement minibatching, for example, not only can change the variance of the distribution by orders of magnitude, but alter the effective loss landscape on which the training unfolds.   However, this sensitivity can be exploited in scenarios where sampling from a precise target distribution is desired, like a posterior in Bayesian machine learning.  To that end, we have shown how the properties of without replacement minibatching enable a new approach (SGWORLD) for stochastic gradient Langevin dynamics with a potential for faster convergence to posterior.

There are a variety of possible avenues for extending our approach in the future.  On a practical level, it would be useful to develop efficient ways to estimate the gradient of the perturbation $\bm\nabla\delta {\cal L}$ for systems with greater numbers of network weights than the simple examples considered here, to allow SGWORLD (and other aspects of the theory) to be tested in real-world applications.  Another natural next step would be generalizing our framework to SGD with momentum and adaptive learning rates, features of algorithms like Adam~\cite{kingma2014adam} commonly used in training large neural networks.   Beyond SGD and its variants, it would be interesting to explore the relevance of stochastic thermodynamics in other learning contexts, particularly biologically-inspired systems. Ref.~\citen{goldt2017stochastic} is an example of this for a model of Hebbian learning.  Along similar lines, one could investigate unsupervised learning algorithms proposed as analogues for biological neural networks~\cite{krotov2019unsupervised}.  Though these do not have the explicit loss landscapes of supervised algorithms, their dynamics could induce effective landscapes over which the network weights evolve.

Finally, deeper understanding of statistical physics analogies often leads to new computational applications.  This has a long history, going back to the development of Monte Carlo techniques at the dawn of the information age, with their close connections to Boltzmann equilibrium~\cite{metropolis1953equation}.  Approximating this equilibrium is at the heart of Bayesian machine learning approaches likes SGLD and SGWORLD.  But could being out of equilibrium have its own benefits, potentially explaining some of the widespread success of SGD?  We have focused on the stationary state, but there is also the critical question of how reliably and quickly we approach this state during the initial phase of training.  Characterizing the inherently nonequilibrium nature of this relaxation process could help us design better training algorithms.

\section*{Code availability}

Code for the calculations described in the text is available at: \url{https://github.com/hincz-lab/ml-nonequilibrium}.

\section*{Acknowledgments}

We acknowledge support from the Scientific and Technological Research Council of Turkey (TUBITAK) under the grant MFAG-119F121 and the KUIS AI Center of Ko\c c University.

\bibliography{ml_ref}

\renewcommand\thesection{Appendix \Alph{section}:}
\renewcommand\thesubsection{\Alph{section}.\Roman{section}}

\setcounter{section}{0}           
\renewcommand{\theequation}{\Alph{section}\arabic{equation}}

\renewcommand\thefigure{\arabic{figure}}
\renewcommand\thetable{\arabic{table}}
\setcounter{figure}{0}
\setcounter{table}{0}

\setcounter{equation}{0}
\section{WR diffusion matrix}

In this appendix we derive the diffusion matrix for the case of SGD with WR minibatching.  Throughout the appendices we will use the Einstein summation notation (unless otherwise noted): repeated indices in a term denote a sum over those indices, with Roman letter indices summing up to $M$, and Greek letter indices summing up to $N$.  As described in Sec.~\ref{sec:sgd_wr}, the SGD update equation for the $a$th parameter $\theta_{t,\alpha}$ at time step $t$ can be expressed using this convention as follows:
\begin{equation}\label{2}
    \theta_{t+1,\alpha} = \theta_{t,\alpha} - \eta\left [\frac{1}{m} \partial_\alpha \ell_i(\bm{\theta}_t)\tilde{e}_{t,i} + \partial_\alpha R(\bm{\theta}_t) \right].
\end{equation}
We use shorthand notation $\partial_\alpha \equiv \partial/\partial \theta_\alpha$ and $\ell_i(\bm\theta) \equiv \ell(\bm{z}_i,\bm\theta)$. $\tilde{e}_{t,i}$ is the $i$th component of a random $M$-dimensional vector chosen at each update step to correspond to the WR minibatching procedure~\cite{hu2017}: we set $m$ distinct entries of the vector $\tilde{\bm{e}}_t$ at random positions to be 1, with the remainder equal to zero.  What makes this procedure WR is that $\tilde{\bm{e}}_{t+1}$ is completely independent from $\tilde{\bm{e}}_t$, and has an identical probability distribution.  Thus a 1 could appear in the same position of the vector in consecutive time steps, meaning the same data point is used in consecutive minibatches.  If the minibatch size $m$ is set equal to $M$ (a minibatch contains the whole data set), then $\tilde{e}_{t,i} = e_i$, where $\bm{e}$ is a vector of all ones.  In this limit the terms in the brackets in Eq.~\eqref{2} equal $\partial_\alpha {\cal L}$, and SGD would reduce to ordinary gradient descent.  When $m < M$, the sum of terms in the brackets differs from the true gradient $\partial_\alpha {\cal L}$, and we can rewrite Eq.~\eqref{2} in the form of Eq.~\eqref{eq_sgd},
\begin{equation}\label{3}
    \theta_{t+1,\alpha} = \theta_{t,\alpha} - \eta\partial_\alpha{\cal L}(\bm{\theta}_t) + \xi_{t,\alpha},
\end{equation}
where the perturbation vector $\bm{\xi}_t$ has components
\begin{equation}\label{4}
\begin{split}
    \xi_{t,\alpha} &= -\frac{\eta}{m} \partial_\alpha \ell_i(\bm{\theta}_t)\tilde{e}_{t,i} + \frac{\eta}{M}\partial_\alpha\ell_i(\bm{\theta}_t)e_i\\
    &= \frac{\eta}{M} \partial_\alpha \ell_i(\bm{\theta}_t)\left(e_i - \frac{M}{m}\tilde{e}_{t,i}\right).
\end{split}
\end{equation}
Because the $\tilde{e}_{t,i}$ at successive update steps are independent, the stochastic dynamics described by Eqs.~\eqref{3}-\eqref{4} is Markovian.

Note that on the first line of Eq.~\eqref{4}, the expression $m^{-1} \partial_\alpha \ell_i \tilde{e}_{t,i}$ in the first term is just a finite sample estimate of $M^{-1} \partial_\alpha \ell_i e_i$ in the second term.  As $m \to M$, this estimate converges to the second term, and hence $\bm{\xi}_t$ becomes small.  In general, when $M \gg m \gg 1$, $\xi_{t,\alpha}$ will have a Gaussian distribution by the central limit theorem, and as mentioned in Sec.~\ref{sec:sgd_wr} we can interpret Eq.~\eqref{4} as an Euler method approximation to the It\^{o} equation, Eq.~\eqref{tr1}, with time units $\Delta t =1$.  The Fokker-Planck equation corresponding to the It\^{o} equation takes the form:
\begin{equation}\label{5}
\begin{split}
    \partial_t P(\bm{\theta},t) &= - \partial_\alpha {\cal J}_\alpha(\bm{\theta},t)\\
\end{split}
\end{equation}
where the components of the $N$ dimensional probability current vector $\bm{\mathcal J}$ are given by
\begin{equation}\label{6}
{\mathcal J}_\alpha(\bm{\theta},t) = -\eta P(\bm{\theta},t) \partial_{\alpha} {\cal L}(\bm{\theta})- \partial_{\beta} \left(D_{\alpha\beta}(\bm{\theta})P(\bm{\theta},t)\right).
\end{equation}
The elements of the $N \times N$ diffusivity matrix $D$ are given by
\begin{equation}\label{7}
D_{\alpha\beta}(\bm{\theta}) = \frac{1}{2}\langle \xi_{t,\alpha} \xi_{t,\beta} \rangle,
\end{equation}
where the brackets denote averaging over all random vectors $\tilde{\bm{e}}_t$ in the definition of $\xi_{t,\alpha}$ in Eq.~\eqref{4}.  To get an explicit expression for the elements $D_{\alpha\beta}(\bm{\theta})$, we use the fact that $\tilde{e}_{t,i}$ has the following moments:
\begin{equation}\label{8}
\begin{split}
 \langle \tilde{e}_{t,i} \rangle &= \frac{m}{M} e_i, \\
 \langle \tilde{e}_{t,i} \tilde{e}_{t,j} \rangle &= \frac{m}{M}\delta_{ij} + \frac{m(m-1)}{M(M-1)} (e_i e_j-\delta_{ij})
\end{split}
\end{equation}
where $\delta_{ij}$ is the Kronecker delta.  Plugging Eq.~\eqref{4} into Eq.~\eqref{7}, and then simplifying using Eq.~\eqref{8}, we find:
\begin{equation}\label{9}
\begin{split}
    &D_{\alpha\beta}(\bm{\theta})\\
    &=\frac{\eta^2}{2M^2} \partial_\alpha \ell_i(\bm{\theta}) \partial_\beta \ell_j(\bm{\theta}) \left\langle \left(e_i - \frac{M}{m}\tilde{e}_{t,i}\right)\left(e_j - \frac{M}{m} \tilde{e}_{t,j}\right)\right\rangle\\
    &= \frac{\eta^2}{2M^2} \partial_\alpha \ell_i(\bm{\theta}) \partial_\beta \ell_j(\bm{\theta}) \left[e_i e_j - \frac{M}{m}\langle \tilde{e}_{t,i} \rangle e_j - \frac{M}{m}e_i \langle \tilde{e}_{t,j}\rangle\right.\\
    &\hspace{0.5\columnwidth} \left.+\frac{M^2}{m^2}\langle \tilde{e}_{t,i}\tilde{e}_{t,j}\rangle  \right]\\
    &=\frac{\eta^2}{2M^2}\frac{M-m}{m(M-1)} \partial_\alpha \ell_i(\bm{\theta}) \partial_\beta \ell_j(\bm{\theta}) ( M\delta_{ij}-e_i e_j).
\end{split}
\end{equation}
If we define the $N\times M$ matrix $V(\bm{\theta})$ with elements
\begin{equation}\label{9b}
V_{\alpha i}(\bm{\theta}) = M^{-1} \partial_\alpha \ell_i(\bm{\theta}),
\end{equation}
then the final result for $D$ can be expressed as shown in Eq.~\eqref{sgd3},
\begin{equation}\label{10}
D(\bm\theta)=\frac{\eta^2(n-1)}{2(M-1)}\left[MV(\bm\theta)V^T(\bm\theta) - \bm{\nabla}L(\bm\theta)\bm{\nabla}L^T(\bm\theta)\right],
\end{equation}
where $(\nabla L(\bm{\theta}))_\alpha = \partial_\alpha L(\bm{\theta}) = M^{-1}\partial_\alpha \ell_i(\bm{\theta}) e_i$.

\setcounter{equation}{0}
\section{WOR diffusion matrix}

To derive the diffusion matrix for WOR minibatching, let us imagine for simplicity that the minibatch size $m$ is chosen so that $n \equiv M/m$ is an integer.  We denote every $n$ update steps an epoch, and we label distinct epochs by $\tau = 1,2,\ldots$.  We use the notation $\bm{\theta}_{\tau t}$ to denote the vector of weights after $t$ update steps within epoch $\tau$, where $t = 0, 1, \ldots, n$.  Components of the vector (or later components of matrices) will be denoted by subscripts following a comma, so the $\alpha$th component of $\bm{\theta}_{\tau t}$ is $\theta_{\tau t,\alpha}$.  Note that $\bm{\theta}_{\tau n} = \bm{\theta}_{(\tau+1) 0}$ is the vector of weights at the beginning of epoch $\tau+1$.  Since we will eventually want to know how the weights change from the beginning of one epoch to the beginning of the next, we also use the shorthand notation $\bm{\theta}_{\tau} \equiv \bm{\theta}_{\tau 0}$ to denote the weights at the beginning of epoch $\tau$.  Within an epoch, the minibatch vectors $\tilde{\bm{e}}_{\tau t}$ again are $M$-dimensional vectors with $m$ random components set to 1, the rest set to zero.  However, they are chosen with no replacement between minibatches in the same epoch, so that within an epoch every one of the $M$ data points is sampled.  In other words,
\begin{equation}\label{14}
\sum_{t=0}^{n-1} \tilde{e}_{\tau t,i} = e_i.
\end{equation}
One can think of the nonzero positions in each successive vector $\tilde{\bm{e}}_{\tau t}$ within an epoch to be successive batches of size $m$ from a random permutation of the integers from $1$ to $M$.  We will assume that the permutation is chosen independently for each epoch.  Thus the set of vectors $\{\tilde{\bm{e}}_{\tau t}\}_{t=0,\ldots,n-1}$ for epoch $\tau$ is independent from the set of vectors $\{\tilde{\bm{e}}_{\tau^\prime t}\}_{t=0,\ldots,n-1}$ for a different epoch $\tau^\prime \ne \tau$.  However, within each epoch the vectors $\tilde{\bm{e}}_{\tau t}$ are not independent of one another, since they have to satisfy Eq.~\eqref{14}.

The analog of Eq.~\eqref{2}, a single update step within an epoch, is:
\begin{equation}\label{15}
\begin{split}
    &\theta_{\tau(t+1),\alpha} \\
    &= \theta_{\tau t,\alpha} - \eta\left [\frac{1}{m} \partial_\alpha \ell_i(\bm{\theta}_{\tau t})\tilde{e}_{\tau t,i} + \partial_\alpha R(\bm{\theta}_{\tau t}) \right], \\
    &\quad 0\le t < n.
\end{split}    
\end{equation}
Within epoch $\tau$, this no longer corresponds to Markovian stochastic dynamics, because the vectors $\bm{e}_{\tau t}$ at different $t$ are not independent.  However the cumulative change over a whole epoch, $\bm{\theta}_{\tau+1} - \bm{\theta}_{\tau}$, will be Markovian, since successive epochs have independent sets of $\bm{e}_{\tau t}$ vectors.  Hence our goal is to construct an epoch-to-epoch Langevin equation that will express $\bm{\theta}_{\tau+1}$ in terms of $\bm{\theta}_{\tau}$.

To facilitate this, let us assume that $\eta$ is small enough that the weight updates within a single epoch $\tau$ will keep us close to the weights at the start of the epoch, $\bm{\theta}_{\tau 0}$.  Then the weight-dependent terms in Eq.~\eqref{15} can be expanded to lowest-order as a Taylor series around $\bm{\theta}_{\tau 0} = \bm{\theta}_{\tau}$, and the equation can be rewritten as
\begin{equation}\label{16}
\begin{split}
&\delta \theta_{\tau (t+1),\alpha}\\ &\approx \delta \theta_{\tau t,\alpha}- \eta\biggl[n \bigl(V_{\alpha i}(\bm{\theta}_{\tau}) + U_{\beta \alpha i}(\bm{\theta}_{\tau}) \delta \theta_{\tau t,\beta}\bigr)\tilde{e}_{\tau t,i}\\&\qquad\qquad\qquad+ X_\alpha(\bm{\theta}_{\tau})+ Y_{\beta\alpha}(\bm{\theta}_{\tau}) \delta \theta_{\tau t,\beta} \biggr].
\end{split}
\end{equation}
Here $\delta \theta_{\tau t,\alpha} \equiv \theta_{\tau t,\alpha} - \theta_{\tau,\alpha}$, the matrix $V$ was defined above in Eq.~\eqref{9b}, and
\begin{equation}\label{17}
\begin{split}
&U_{\beta\alpha i}(\bm{\theta}) \equiv M^{-1} \partial_\beta \partial_\alpha \ell_i (\bm{\theta}),\\ 
&X_\alpha(\bm{\theta}) \equiv \partial_\alpha R(\bm{\theta}),\\  &Y_{\beta\alpha}(\bm{\theta}) \equiv \partial_\beta \partial_\alpha R(\bm{\theta}).
\end{split}
\end{equation}
If we additionally define an $N \times N$ matrix $Q_{\tau t}$ and $N$-dimensional vector $\bm{u}_{\tau t}$ by
\begin{equation}\label{18}
\begin{split}
    &Q_{\tau t,\beta\alpha} \equiv n U_{\beta \alpha i}(\bm{\theta}_{\tau}) \tilde{e}_{\tau t,i} + Y_{\beta\alpha}(\bm{\theta}_{\tau}),\\
    &u_{\tau t,\alpha} \equiv n V_{\alpha i}(\bm{\theta}_{\tau}) \tilde{e}_{\tau t,i} + X_{\alpha}(\bm{\theta}_{\tau}),
\end{split}
\end{equation}
then Eq.~\eqref{16} can be expressed as
\begin{equation}\label{19}
    \delta \bm{\theta}_{\tau(t+1)} = (I - \eta Q_{\tau t})\delta \bm{\theta}_{\tau t} - \eta \bm{u}_{\tau t},
\end{equation}
where $I$ is an $N\times N$ identity matrix.  Eq.~\eqref{19} can be iterated over the $n$ update steps in an epoch, yielding
\begin{equation}\label{20}
\bm{\theta}_{\tau+1} - \bm{\theta}_{\tau} = \delta\bm{\theta}_{\tau n} \approx  -\eta\sum_{t=0}^{n-1} \left[\prod_{t^\prime = t+1}^{n-1} (I - \eta Q_{\tau t^\prime}) \right] \bm{u}_{\tau t}. 
\end{equation}
We will expand the product in the square brackets to first order in $\eta$, giving $\prod_{t^\prime = t+1}^{n-1} (I - \eta Q_{\tau t^\prime}) \approx I - \eta \sum_{t^\prime=t+1}^{n-1} Q_{\tau t^\prime}$, which allows us to write Eq.~\eqref{20} as
\begin{equation}\label{21}
\bm{\theta}_{\tau+1} - \bm{\theta}_{\tau}  \approx  -\eta\sum_{t=0}^{n-1} \bm{u}_{\tau t} + \eta^2 \sum_{t=0}^{n-1}\sum_{t^\prime = t+1}^{n-1} Q_{\tau t^\prime}\bm{u}_{\tau t}.
\end{equation}
By plugging in Eq.~\eqref{18} and rearranging terms (using Eq.~\eqref{14} to simplify), we can rewrite Eq.~\eqref{21} in the form of a Langevin equation (analogous to Eq.~\eqref{3}),
\begin{equation}\label{22}
    \theta_{\tau+1,\alpha} \approx \theta_{\tau,\alpha} -\eta \partial_\alpha \hat{\cal L}(\bm{\theta}_{\tau}) + \hat{\xi}_{\tau,\alpha},
\end{equation}
Here the effective loss function $\hat{\cal{L}}(\bm{\theta})$ is given by:
\begin{equation}\label{23}
\begin{split}
&\hat{\cal{L}}(\bm{\theta}) = n {\cal L}(\bm{\theta})
- \frac{\eta n(n-1)}{4}\bigg[\partial_\alpha {\cal L}(\bm{\theta}) \partial_\alpha {\cal L}(\bm{\theta})\\
&+ \frac{\partial_\alpha L(\bm{\theta})\partial_\alpha L(\bm{\theta})}{M-1}- \frac{\partial_\alpha \ell_i(\bm{\theta}) \partial_\alpha \ell_i(\bm{\theta})}{M(M-1)} \bigg],
\end{split}
\end{equation}
and the random displacement $\hat{\xi}_{\tau,\alpha}$ has the form
\begin{equation}\label{24}
    \hat{\xi}_{\tau,\alpha} = \eta^2 \left[Z_{\alpha i}(\bm{\theta}_{\tau}) \zeta_{\tau,i} + S_{\alpha i j}(\bm{\theta}_{\tau})\chi_{\tau,ij}\right],
\end{equation}
where
\begin{equation}\label{25}
\begin{split}
    &Z_{\alpha i}(\bm{\theta}) \equiv n Y_{\alpha\beta}(\bm{\theta})V_{\beta i}(\bm{\theta}) - n U_{\alpha\beta i}(\bm{\theta})X_\beta(\bm{\theta}),\\
    &S_{\alpha i j}(\bm{\theta}) \equiv n^2 U_{\alpha \beta i}(\bm{\theta})V_{\beta j}(\bm{\theta}).
\end{split}
\end{equation}
The quantities $\zeta_{\tau,i}$ and $\chi_{\tau,ij}$ in Eq.~\eqref{24} are stochastic functions of the minibatches for epoch $\tau$,
\begin{equation}\label{26}
\begin{split}
    \zeta_{\tau,i} &= -\frac{n-1}{2}+ \sum_{t=0}^{n-1} (n-1-t)\tilde{e}_{\tau t,i},\\
    \chi_{\tau,ij} &= -\frac{(M-m)(1-\delta_{ij})}{2(M-1)} + \sum_{t=0}^{n-1}\sum_{t^\prime= t+1}^{n-1} \tilde{e}_{\tau t^\prime,i} \tilde{e}_{\tau t,j}.
    \end{split}
\end{equation}
Their first moments are zero, $\langle \zeta_{\tau,i}\rangle = \langle \chi_{\tau,ij} \rangle =0$, when averaged over all possible sets of minibatches for an epoch, and hence $\langle \hat{\xi}_{\tau,\alpha}\rangle =0$.  The second moments are given by the following expressions (note we do not sum over repeated indices in these expressions):
\begin{equation}\label{27}
    \begin{split}
        &\langle \zeta_{\tau,i} \zeta_{\tau,j}\rangle = a_0(m,M) \delta_{ij},\\         
        &\langle \zeta_{\tau,i} \chi_{\tau,jk}\rangle = a_1(m,M)(1-\delta_{jk})(\delta_{ik}-\delta_{ij}),\\
        &\langle \chi_{\tau,ij} \chi_{\tau,kl}\rangle\\ &= (1-\delta_{ij})(1-\delta_{kl})\biggl[a_2(m,M)\bigl[\delta_{ik}(1-\delta_{jl})\\ &+\delta_{jl}(1-\delta_{ik})\bigr] 
         + a_3(m,M) \delta_{ik}\delta_{jl}+ a_4(m,M) \delta_{il}\delta_{jk}\\
        &\qquad\qquad + a_5(m,M)\bigl[\delta_{il}(1-\delta_{jk})+\delta_{jk}(1-\delta_{il})\bigr]\biggr],
    \end{split}
\end{equation}
where
\begin{equation}\label{28}
\begin{split}
    &a_0(m,M) \equiv \frac{(M-m)(M+m)}{12 m^2},\\
    &a_1(m,M) \equiv \frac{(n+1)(M-m)}{12(M-1)},\\
    &a_2(m,M) \equiv \frac{(M-m)(M(M-4)+(M-4)m+6)}{12(M-2)(M-1)^2},\\
    &a_3(m,M) \equiv \frac{(M-m)(M+m-2)}{4(M-1)^2},\\ 
    &a_4(m,M) \equiv -\frac{(M-m)^2}{4(M-1)^2},\\
    &a_5(m,M) \\
    &\quad \equiv -\frac{(M-m)(12M+(M-4)(M^2+(6+M)m))}{12(M-2)(M-1)^2 M}.
\end{split}
\end{equation}
As in the previous section, the Langevin equation in Eq.~\eqref{22} corresponds to a Fokker-Planck equation for the weight probability density $P(\bm{\theta},\tau)$,
\begin{equation}\label{29}
\begin{split}
    \partial_\tau P(\bm{\theta},\tau) &= - \partial_\alpha \hat{{\mathcal J}}_\alpha(\bm{\theta},\tau)\\
\end{split}
\end{equation}
with a probability current vector components $\hat{\mathcal J}_\alpha(\bm{\theta},\tau)$ given by
\begin{equation}\label{30}
\hat{\mathcal J}_\alpha(\bm{\theta},\tau) = -\eta P(\bm{\theta},\tau) \partial_{\alpha} {\cal \hat{L}}(\bm{\theta})- \partial_{\beta} \left(\hat{D}_{\alpha\beta}(\bm{\theta})P(\bm{\theta},\tau)\right).
\end{equation}
Using the moment results from Eq.~\eqref{27}, the diffusivity matrix $\hat{D}_{\alpha\beta}(\bm{\theta}) = \langle \hat{\xi}_{\tau,\alpha} \hat{\xi}_{\tau,\beta}\rangle/2$ can be expressed in the following way (hiding functional dependences for simplicity):
\begin{equation}\label{31}
\begin{split}
&\hat{D} = \frac{1}{2}\eta^4\biggl[a_0 ZZ^T + a_1 \bigl(Z(B^T-C^T) + (B - C)Z^T \bigr)\\ &+ a_2 \bigl(BB^T +CC^T  \bigr)
+ (a_3-2a_2) F \\&+ (a_4-2a_5)G + a_5 \bigl(BC^T +CB^T  \bigr) \biggr].
\end{split}
\end{equation}
To define the matrices $B$, $C$, $F$, and $G$ in the above expression, let us introduce the tensor $S^\Delta_{\alpha ij}(\bm{\theta})$ through
\begin{equation}\label{32}
S^\Delta_{\alpha i j}(\bm{\theta}) = \begin{cases} S_{\alpha i j}(\bm{\theta}) & i \ne j\\
0 & i=j
\end{cases}.
\end{equation}
Then
\begin{equation}\label{33}
\begin{split}
B_{\alpha i}(\bm{\theta}) &\equiv S^\Delta_{\alpha j i}(\bm{\theta})e_j,\\
C_{\alpha i}(\bm{\theta}) &\equiv S^\Delta_{\alpha i j}(\bm{\theta})e_j,\\
F_{\alpha\beta}(\bm{\theta}) &\equiv S^\Delta_{\alpha i j}(\bm{\theta})S^\Delta_{\beta i j}(\bm{\theta}),\\
G_{\alpha\beta}(\bm{\theta}) &\equiv S^\Delta_{\alpha i j}(\bm{\theta})S^\Delta_{\beta j i}(\bm{\theta}).
\end{split}
\end{equation}
Since all the prefactors $a_i(m,M) = 0$ for $m=M$, we see that $\hat{D}$ vanishes in the deterministic limit.  Additionally, the structure of Eqs.~\eqref{31}-\eqref{33} makes it clear the $\hat{D}$ is symmetric.  In all the numerical examples explored so far, Eq.~\eqref{31} can be approximated very well by a single term (since the other ones are relatively small),
\begin{equation}
    \hat{D} \approx \frac{1}{2}\eta^4 a_2 B B^T.
\end{equation}

\setcounter{equation}{0}
\section{Stationary state probability distribution near a local minimum of the loss function}

The Fokker-Planck equations for the two SGD types, Eqs.~\eqref{5}-\eqref{6} and \eqref{29}-\eqref{30} respectively, share the same form, except in the second case ${\cal L}$, $D$, $t$ are replaced by $\hat{\cal L}$, $\hat{D}$, $\tau$.  We will formulate the discussion in this section in terms of the WR quantities ${\cal L}$, $D$, $t$, but all the following results generalize to WOR with the above replacements.

When we take the limit $t\to\infty$ the system will go to a stationary state with weight probability distribution $P(\bm{\theta},t) \to P^\text{s}(\bm{\theta})$.  The form of this stationary distribution is complex in general, but there is one common case where it simplifies greatly:  if the dynamics at large $t$ are just small fluctuations around the global minimum of the regularized loss function ${\cal L}(\bm{\theta})$.  Alternatively we might imagine a case where for a long stretch of training the dynamics is trapped near a deep local minimum, and hence the probability distribution approaches a quasistationary distribution centered at that minimum.  In that case the analysis below is also approximately valid.

Let $\bm{\theta}_0$ be the position of a global (or local) minimum of ${\cal L}(\bm{\theta})$.  Hence $\partial_\alpha {\cal L}(\bm{\theta}_0) = 0$, and the associated Hessian matrix $H_0$ is given by
\begin{equation}
    H_{0,\alpha\beta} = \partial_\alpha \partial_\beta {\cal L}(\bm{\theta}_0).
\end{equation}
The probability current, Eq.~\eqref{6}, can be 
approximated in the vicinity of the minimum by Taylor expanding ${\cal L}(\bm{\theta})$ and $D(\bm{\theta})$, yielding
\begin{equation}\label{36}
\begin{split}
    &{\mathcal J}_\alpha(\bm{\theta},t)\\
    &\qquad \approx -\eta P(\bm{\theta},t) H_{0,\alpha\beta}(\theta_\beta - \theta_{0,\beta}) - D_{0,\alpha\beta} \partial_\beta P(\bm{\theta},t).
    \end{split}
\end{equation}
Note we have kept only the zeroth order contribution to diffusivity, $D_0 \equiv D(\bm{\theta}_0)$, since this determines the stationary distribution to leading order in $\eta$.  The correction introduced by keeping higher-order Taylor series terms for $D(\bm{\theta})$, accounting for the position-dependence of the diffusivity near $\bm{\theta}_0$, is shown in Appendix D. It is convenient to switch variables to $\delta \bm{\theta} = \bm{\theta}-\bm{\theta}_0$, so that Eq.~\eqref{36} takes the form
\begin{equation}\label{37}
{\mathcal J}_\alpha(\delta \bm{\theta},t) \approx -\eta P(\delta \bm{\theta},t) H_{0,\alpha\beta}\delta \theta_\beta - D_{0,\alpha\beta} \partial_\beta P(\delta \bm{\theta},t),
\end{equation}
and the associated Fokker-Planck equation is just $\partial_t P(\delta \bm{\theta},t) = -\partial_\alpha {\mathcal J}_\alpha(\delta \bm{\theta},t)$.

Before looking at the stationary distribution near the minimum, it is instructive to consider the dynamical equations for the first and second moments of the weight vector components, assuming dynamics described by Eq.~\eqref{37}.  The moments are:
\begin{equation}\label{38}
\begin{split}
\mu_\alpha&=\int d(\delta\bm{\theta})\, \delta \theta_\alpha P(\delta \bm{\theta},t),\\
\Sigma_{\alpha\beta} &= \int d(\delta\bm{\theta})\, \delta \theta_\alpha \delta \theta_\beta P(\delta \bm{\theta},t).
\end{split}
\end{equation}
The first moment obeys the dynamical equation
\begin{equation}\label{39}
\begin{split}
&\frac{d\mu_\alpha}{dt}\\
&= \int d(\delta\bm{\theta})\, \delta \theta_\alpha \partial_tP(\delta \bm{\theta},t)\\
&= -\int d(\delta\bm{\theta})\, \delta \theta_\alpha \partial_\beta {\mathcal J}_\beta(\delta \bm{\theta},t)\\
&= \int d(\delta\bm{\theta})\, {\mathcal J}_\alpha(\delta \bm{\theta},t)\\
&= -\int d(\delta\bm{\theta})\, \left[ \eta P(\delta \bm{\theta},t) H_{0,\alpha\beta}\delta \theta_\beta + D_{0,\alpha\beta} \partial_\beta P(\delta \bm{\theta},t)\right]\\
&=-\eta H_{0,\alpha\beta}\mu_\beta.
\end{split}
\end{equation}
In the fourth line we have used integration by parts, with the boundary terms negligible assuming the probability distribution rapidly decays to zero at large $\delta \bm{\theta}$.  Similar considerations allow us to neglect the integral over $\partial_\beta P$ in the sixth line.  If $P(\delta \bm{\theta},t) \to P^s(\delta \bm{\theta})$ as $t \to \infty$, then $d\mu_\alpha/dt \to 0$.  Since the Hessian matrix at a local minimum is positive definite, the only solution to Eq.~\eqref{39} at large $t$ is $\bm{\mu} =0$.  Hence the mean of the stationary distribution $P^s(\delta \bm{\theta})$ is at the origin, corresponding to the mean weight vector being $\bm{\theta}_0$.

A similar calculation for the second moment $\Sigma_{\alpha\beta}$ gives
\begin{equation}\label{40}
\begin{split}
\frac{d\Sigma_{\alpha\beta}}{dt}&= \int d(\delta\bm{\theta})\, \delta \theta_\alpha\delta \theta_\beta \partial_tP(\delta \bm{\theta},t)\\
&= -\int d(\delta\bm{\theta})\, \delta \theta_\alpha\delta \theta_\beta \partial_\gamma {\mathcal J}_\gamma(\delta \bm{\theta},t)\\
&= \int d(\delta\bm{\theta})\, (\delta \theta_\beta {\mathcal J}_\alpha(\delta \bm{\theta},t) + \delta \theta_\alpha {\mathcal J}_\beta(\delta \bm{\theta},t))\\
&= -\int d(\delta\bm{\theta})\, \bigg[ \eta P(\delta \bm{\theta},t) (H_{0,\alpha\gamma}\delta \theta_\gamma\delta \theta_\beta\\
&\qquad\quad +H_{0,\beta\gamma}\delta \theta_\gamma\delta \theta_\alpha)+ \delta \theta_\beta D_{0,\alpha\gamma} \partial_\gamma P(\delta \bm{\theta},t)\\
&\qquad\quad+ \delta \theta_\alpha D_{0,\beta\gamma} \partial_\gamma P(\delta \bm{\theta},t) \bigg]\\
&= -\int d(\delta\bm{\theta})\, \bigg[ \eta P(\delta \bm{\theta},t) (H_{0,\alpha\gamma}\delta \theta_\gamma\delta \theta_\beta\\
&\quad +H_{0,\beta\gamma}\delta \theta_\gamma\delta \theta_\alpha)- (D_{0,\alpha\beta}+D_{0,\beta\alpha}) P(\delta \bm{\theta},t)\bigg]\\
&=-\eta H_{0,\alpha \gamma} \Sigma_{\gamma\beta} -\eta \Sigma_{\alpha\gamma} H_{0,\gamma\beta} + 2 D_{0,\alpha\beta},
\end{split}
\end{equation}
where we have used the fact that $D_0$, $H_0$, and $\Sigma$ are all symmetric matrices.  At stationarity, when the left-hand side of Eq.~\eqref{40} is zero, we see the following condition must be satisfied, the Lyapunov relation of Eq.~\eqref{n3},
\begin{equation}\label{41}
    H_0\Sigma + \Sigma H_0 = 2\eta^{-1}D_0.
\end{equation}

Given that $\bm{\mu} = 0$ and $\Sigma$ satisfies Eq.~\eqref{41}, one can posit a Gaussian form for the stationary distribution,
\begin{equation}\label{42}
    P^s(\delta \bm{\theta}) = \frac{1}{\sqrt{(2\pi)^N \det \Sigma}} \exp\left(-\frac{1}{2}\delta \theta_\alpha \Sigma^{-1}_{\alpha\beta} \delta \theta_\beta \right).
\end{equation}
To verify that this satisfies the Fokker-Planck equation, one can calculate the corresponding stationary probability current
\begin{equation}\label{43}
\begin{split}
    {\mathcal J}_\alpha^s(\delta \bm{\theta}) &= -\eta P^s(\delta \bm{\theta}) H_{0,\alpha\beta}\delta \theta_\beta - D_{0,\alpha\beta} \partial_\beta P^s(\delta \bm{\theta})\\
    &= -\eta P^s(\delta \bm{\theta}) H_{0,\alpha\beta}\delta \theta_\beta - D_{0,\alpha\beta} \partial_\beta P^s(\delta \bm{\theta})\\
    &= \left(\eta  H_{0,\alpha\gamma} \Sigma_{\gamma\beta} - D_{0,\alpha\beta}\right) \partial_\beta P^s(\delta \bm{\theta}),
\end{split}
\end{equation}
and then check that its divergence is zero.  We write the divergence of the current as follows
\begin{equation}\label{44}
\begin{split}
    \partial_\alpha {\mathcal J}_\alpha^s(\delta \bm{\theta}) &= \left(\eta  H_{0,\alpha\gamma} \Sigma_{\gamma\beta} - D_{0,\alpha\beta}\right) \partial_\alpha \partial_\beta P^s(\delta \bm{\theta})\\
    &=\text{tr}\left[\left(\eta H_0 \Sigma -D_0\right){\cal K}(\delta \bm{\theta}) \right],
\end{split}
\end{equation}
where ${\cal K}_{\alpha\beta}(\delta \bm{\theta}) \equiv \partial_\alpha\partial_\beta P^s(\delta \bm{\theta})$ is a symmetric matrix.  Eq.~\eqref{44} can be rearranged as
\begin{equation}\label{45}
\begin{split}
    \partial_\alpha {\mathcal J}_\alpha^s(\delta \bm{\theta})
    &=\text{tr}\left[\left(\frac{1}{2}\eta (H_0 \Sigma +\Sigma H_0)\right.\right.\\
    &\qquad\quad\left.\left.+ \frac{1}{2}\eta(H_0\Sigma - \Sigma H_0)-D_0\right){\cal K}(\delta \bm{\theta}) \right]\\
&=\text{tr}\left[\left(\frac{1}{2}\eta (H_0 \Sigma +\Sigma H_0) -D_0\right){\cal K}(\delta \bm{\theta}) \right],   
\end{split}
\end{equation}
where the third line follows from the fact that the product of an antisymmetric and symmetric matrix has zero trace.  The terms in the brackets above sum to zero because of Eq.~\eqref{41}, and hence the divergence of the current is zero.

If we write the stationary probability current as $\bm{J}^s(\delta \bm{\theta}) = \bm{v}^s(\delta \bm{\theta}) P^s(\delta \bm{\theta})$, defining a velocity field $\bm{v}(\delta \bm{\theta})$, then from Eq.~\eqref{43} it follows that 
\begin{equation}
\begin{split}
    \bm{v}^s(\delta \bm{\theta}) &= -(\eta H_0 \Sigma -D_0)\Sigma^{-1}\delta \bm{\theta} \equiv -C\Sigma^{-1}\delta \bm\theta,
\end{split}
\end{equation}
defining the circulation matrix of Eq.~\eqref{n5}.

\setcounter{equation}{0}
\section{Correction to the stationary state properties near a local minimum due to a varying diffusion matrix}

In Sec.~\ref{sec:sgworld}, we used a generalized version of the stationary state theory described in Appendix C.  The starting premise is the same: we are looking at the small $\eta$ limit of the stationary distribution near a local minimum $\bm\theta_0$, where it is approximately Gaussian, as given by Eq.~\eqref{42}.  However, instead of approximating $D(\bm\theta)$ by a constant matrix $D(\bm\theta_0)$, we expand $D(\bm\theta)$ as a Taylor series up to second order:
\begin{equation}\label{d1}
\begin{split}
D_{\alpha\beta}(\bm\theta) &\approx D(\bm\theta_0) + \partial_\gamma D_{\alpha\beta}(\bm\theta_0)\delta \theta_\gamma \\
&\qquad + \frac{1}{2} \partial_\mu\partial_\gamma D_{\alpha\beta}(\bm\theta_0) \delta \theta_\mu \delta\theta_\gamma.
\end{split}
\end{equation}
The calculation proceeds analogously to the one in Appendix C, but with Eq.~\eqref{d1} substituted for $D_0$ in Eq.~\eqref{36}.  The final form of Eq.~\eqref{39} for the first moment remains unchanged, so in the stationary state limit $\bm{\mu}=0$, and hence the mean of the stationary distribution stays at $\bm\theta_0$.

However the relation for the second moment, Eq.~\eqref{40}, gets modified by a correction.  In the $t \to \infty$ limit, where $d\Sigma_{\alpha\beta}/dt \to 0$, the condition on $\Sigma$ now takes the form:
\begin{equation}\label{d2}
\begin{split}
0 &= -\eta H_{0,\alpha \gamma} \Sigma_{\gamma\beta} -\eta \Sigma_{\alpha\gamma} H_{0,\gamma\beta}\\
&\qquad + 2 \int d(\delta\bm\theta)\, D_{\alpha\beta}(\bm\theta) P^s(\delta\bm\theta)\\
&= -\eta H_{0,\alpha \gamma} \Sigma_{\gamma\beta} -\eta \Sigma_{\alpha\gamma} H_{0,\gamma\beta}\\
&\qquad + 2 D_{\alpha\beta}(\bm\theta_0) + \partial_\mu\partial_\gamma D_{\alpha\beta}(\bm\theta_0) \Sigma_{\mu\gamma}.
\end{split}
\end{equation}
We can think of $\partial_\mu\partial_\gamma D_{\alpha\beta}(\bm\theta_0)$ as components ${\cal H}_{D,\alpha\beta\mu\gamma}(\bm\theta_0)$ of a fourth-order Hessian-like tensor ${\cal H}_D(\bm\theta_0)$ for the diffusion matrix.  The last term in Eq.~\eqref{d2} is a contraction of this tensor with the $\Sigma$ matrix, yielding a matrix with components $({\cal H}_D(\bm\theta_0) \Sigma)_{\alpha\beta} \equiv \partial_\mu\partial_\gamma D_{\alpha\beta}(\bm\theta_0) \Sigma_{\mu\gamma}$.  Thus Eq.~\eqref{d2} can be written as:
\begin{equation}
    \label{d3}
    H_0 \Sigma +\Sigma H_0 = \eta^{-1} (2 D(\bm\theta_0) + {\cal H}_{D}(\bm\theta_0) \Sigma),
\end{equation}
generalizing Eq.~\eqref{41}.
\end{document}